%% file: ITS.tex
\documentclass[lettersize,journal]{IEEEtran}
\usepackage{amsmath,amsfonts}
\usepackage{array}
\usepackage{textcomp}
\usepackage{stfloats}
\usepackage{url}
\usepackage{verbatim}
\usepackage{graphicx}
\usepackage{cite}

\usepackage{pifont}
\usepackage{amssymb}
\usepackage{booktabs,multirow}
\usepackage{epstopdf}
\usepackage{makecell} 
\usepackage[linewidth=1pt]{mdframed}
\usepackage[export]{adjustbox}
\usepackage{arydshln}

\usepackage{hhline}
\usepackage{diagbox}
\usepackage{color}
\usepackage{hyperref}
\hypersetup{colorlinks=true,
linkcolor=blue,
anchorcolor=blue,
citecolor=blue}

\usepackage{tikz}
\definecolor{lime}{HTML}{A6CE39}
\DeclareRobustCommand{\orcidicon}{%
    \begin{tikzpicture}
    \draw[lime, fill=lime] (0,0) 
    circle [radius=0.16] 
    node[white] {{\fontfamily{qag}\selectfont \tiny ID}};    \draw[white, fill=white] (-0.0625,0.095) 
    circle [radius=0.007];    \end{tikzpicture}
    \hspace{-2mm}}
\foreach \x in {A, ..., Z}{%
    \expandafter\xdef\csname orcid\x\endcsname{\noexpand\href{https://orcid.org/\csname orcidauthor\x\endcsname}{\noexpand\orcidicon}}
    }

\begin{document}

\title{Enhancing Traffic Object Detection in Variable Illumination with RGB-Event Fusion}

\author{{Zhanwen~Liu$^{\ast}$\orcidA{},~Nan~Yang$^{\ast}$\orcidB{},~Yang~Wang$^{\dagger}$\orcidC{},~Yuke~Li,~Xiangmo~Zhao,~Fei-Yue~Wang}
\thanks{$^{\ast}$Co-first author. $^{\dagger}$Corresponding author.}
\thanks{Zhanwen Liu, Nan Yang, Yang Wang and Xiangmo Zhao are with the School of Information Engineering, Chang'an University, Shaanxi, Xi’an 710000, China (e-mail: zwliu@chd.edu.cn; 2022024001@chd.edu.cn; ywang120@chd.edu.cn; xmzhao@chd.edu.cn).}
\thanks{Yuke Li is with the Waytous Co. Ltd., Beijing 100083, China (e-mail: liyuke14@mails.ucas.ac.cn).}
\thanks{Fei-Yue Wang is with the Institute of Automation, Chinese Academy of Sciences, Beijing 100083, China (e-mail: feiyue.wang@ia.ac.cn).}}

\maketitle
\begin{abstract}
Traffic object detection under variable illumination is challenging due to the information loss caused by the limited dynamic range of conventional frame-based cameras. To address this issue, we introduce bio-inspired event cameras and propose a novel Structure-aware Fusion Network (SFNet) that extracts sharp and complete object structures from the event stream to compensate for the lost information in images through cross-modality fusion, enabling the network to obtain illumination-robust representations for traffic object detection. Specifically, to mitigate the sparsity or blurriness issues arising from diverse motion states of traffic objects in fixed-interval event sampling methods, we propose the Reliable Structure Generation Network (RSGNet) to generate Speed Invariant Frames (SIF), ensuring the integrity and sharpness of object structures. Next, we design a novel Adaptive Feature Complement Module (AFCM) which guides the adaptive fusion of two modality features to compensate for the information loss in the images by perceiving the global lightness distribution of the images, thereby generating illumination-robust representations. Finally, considering the lack of large-scale and high-quality annotations in the existing event-based object detection datasets, we build a DSEC-Det dataset, which consists of 53 sequences with 63,931 images and more than 208,000 labels for 8 classes. Extensive experimental results demonstrate that our proposed SFNet can overcome the perceptual boundaries of conventional cameras and outperform the frame-based methods, \emph{e.g.}, YOLOX by 7.9\% in mAP50 and 3.8\% in mAP50:95. Our code and dataset will be available at \url{https://github.com/YN-Yang/SFNet}.
\end{abstract}

\section{Introduction}
\IEEEPARstart{T}{raffic} object detection aims to accurately 
recognize and locate traffic objects, making it a critical 
component of autonomous driving systems \cite{wang2023foundation,liu2021cascade,valiente2023robust,mia2024secure,liu2024boosting,liu2020sadanet,guo2023sustainability} and serving as the 
foundation for downstream tasks such as object tracking and 
trajectory 
prediction\cite{TENG,10122127,9965619,li2024intention,li2023regional}. Notably, the success of existing traffic object detection methods is heavily contingent 
upon image quality\cite{tian2013rear,zhang2021c2fda,liu2020scale,chen2021deep}. However, the limited capacitance capacity in 
the integral imaging circuit of frame-based cameras restricts 
their dynamic range \cite{4444573}, making it difficult to 
achieve stable imaging in poor lighting conditions (\emph{e.g.}, 
low-light and over-exposure), which will result in decreased 
image contrast and information loss and hinder the extraction of 
representative features for traffic object detection\cite{liu2023multi}.

\input{PDF/Figure/Fig_FeatureAfterFusion}

To compensate for the limitations of frame-based cameras, we propose to introduce the bio-inspired event cameras \cite{brebion2021real,munir2021ldnet,chen2022progressivemotionseg,trauth2023toward,liu2024enhancing,tan2022multi,yang2023joint,jha2023driver,li2024event,li2023emergent} for traffic object detection. Unlike frame-based cameras, event cameras detect changes in light intensity at the pixel level and asynchronously output the event stream. With high dynamic range (\textgreater 120 dB), event cameras exhibit remarkable stability and robustness in poor lighting conditions, presenting great potential in traffic object detection under variable illumination conditions. However, the event stream captured by event cameras lacks color and fine-grained texture information, which is very important for high-performance detection\cite{wang2023visevent}. Therefore, to achieve robust traffic object detection under poor lighting conditions, we propose leveraging the complementary properties of both event and image modalities.

To achieve this goal, two challenges must be comprehensively considered and addressed. The first challenge is to overcome the modality differences between events and images and extract high-quality structural information for diverse motion states from the event stream. To this end, a typical solution for existing methods \cite{maqueda2018event,zhu2018ev,park2016performance,lagorce2016hots,sironi2018hats,kim2021n,zihao2018unsupervised,zhu2019unsupervised} resort to using a fixed time window to sample and compress temporal information of the event stream, generating image-like tensors for subsequent feature extraction. However, the fixed event sampling is susceptible to introducing sparsity or blurriness issues for moving objects. As shown in Fig. \ref{fig2}(b, c), when the time window is small, slow-moving objects will exhibit sparse structures that can be easily confused with noise \cite{wan2022s2n, duan2024led}; when the time window is large, fast-moving objects will show blurry structures that inhibit the structure perceiving\cite{sun2019fab}. Such low-quality structural information introduces uncertainties in generated features. Another category of methods, known as motion compensation \cite{stoffregen2019event, gallego2019focus, xu2020robust, stoffregen2019event2, shiba2022event,gallego2018unifying, hagenaars2021self, gu2021spatio, paredes2023taming,shiba2022secrets,gehrig2021raft,Wu_2024_ICRA}, estimate the motion of the event stream to compensate the events to a reference timestamp, producing images of warped events (IWE). The generated IWE typically exhibits sharp edges and is widely used as input for downstream tasks. However, the existing motion compensation methods mainly rely on two assumptions for stable operation, the constant illumination assumption and the linear motion assumption. In autonomous driving scenarios, changing lighting conditions (\emph{e.g.}, low-light, and headlight glare) and 
diverse motion states can easily break these two assumptions\cite{shiba2022secrets, paredes2023taming}, leading to motion estimation failures. Therefore, a light-robust and motion-robust representation method is required to ensure that objects with complex and variable motion patterns can obtain complete and sharp structural information.

Further, the poor light conditions will result in non-uniform information loss and contrast degradation across the image. Therefore, another challenge lies in effectively perceiving the degree of information loss and contrast reduction across the image and executing adaptive information compensation. Currently, various fusion methods have been proposed for object detection, which can be categorized into two strategies: late fusion and middle fusion. Late fusion methods \cite{chen2018pseudo,jiang2019mixed,wang2023drive,li2019event} directly combine the detection results of two modalities, but lack interaction between different modalities. Middle fusion methods \cite{liu2021attention,cao2021fusion,wang2023vehicle,tomy2022fusing,zhou2022rgb} achieve fusion by simple concatenation and attention mechanisms during the feature extraction stage. However, these methods cannot effectively perceive the degrees of information loss across the images and fail to adaptively complement information, which is essential for object detection in variable illumination conditions.

\input{PDF/Figure/Fig_FixedTimeWindows}

To address the above issues, we propose a novel Structure-aware Fusion Network (SFNet), which consists of two cascaded steps: Reliable Structure Generation Network (RSGNet) and Adaptive Feature Complement Module (AFCM), accounting for motion-robust high-quality structural information generation and illumination-robust representations modeling through adaptive cross-modality fusion, respectively, as shown in Fig. \ref{fig1}. Specifically, first, the RSGNet is devised to generate Speed Invariant Frames (SIF) by using a large time window to aggregate complete structural information for slow-moving objects and generating sharp edges for fast-moving objects with the assistance of motion cues extracted from event stream, capable of handling diverse motion patterns. The resulting SIF is motion-robust and provides complementary structural cues for images. Second, the AFCM is designed to perceive the lightness distribution in images by calculating the global spatial correlation between image and event modalities, which is used to guide the adaptive feature fusion between image and event modalities. Besides, considering the impact of noise in the event stream on the fusion process, we leverage the local smoothness of the image modality to suppress event noise. This adaptive fusion process allows us to leverage the strengths of both modalities and tailor the feature fusion to information loss of each region in images. Finally, to facilitate further research on RGB-Event fusion-based object detection, we provide rich and accurate object detection annotations on the DSEC dataset \cite{gehrig2021dsec} and propose the DSEC-Det dataset.

In summary, the main contributions of this paper are as follows:

(1) We propose a RSGNet, which generates Speed Invariant Frames to ensure the integrity and sharpness of structures for objects with diverse motion patterns.

(2) We design an AFCM, which estimates information loss by perceiving the lightness distribution of the image modality and adaptively extracts event features to complement it.

(3) We build a large-scale RGB-Event object detection dataset DSEC-Det, comprising accurate object detection
annotations and diverse driving scenes with variable illumination conditions. 

(4) We benchmark various state-of-the-art object detection methods on our dataset and demonstrate that 
our SFNet significantly outperforms existing methods.

\section{Related Work}

This section first provides a comprehensive overview of the event-based object detection datasets. Subsequently, we review existing event-based object detection methods and event representation methods.

\subsection{Datasets}

\textbf{Event modality.} The datasets in the event modality include the Gen1 dataset \cite{de2020large} and the 1 Mpx dataset \cite{perot2020learning}, both of which contain substantial data. Gen1 includes over 255,000 labeled cars and pedestrians with a resolution of 304 × 240 and annotated frequencies of 1 Hz, 2 Hz, or 4 Hz. 1 Mpx includes over 25 million labeled pedestrians, two wheelers, cars, trucks, buses, traffic signs, and traffic lights with a higher resolution of 1280 × 720 and an annotated frequency of 60 Hz. However, both datasets exclusively contain event modality data and not publicly released image modality data.

\textbf{RGB\&Event modality.} EventKITTI\cite{liang2022global} utilizes the ESIM\cite{rebecq2018esim} simulator to generate event data for the KITTI \cite{geiger2012we} dataset, yet the simulated events significantly differ from real events. Some works\cite{tomy2022fusing}\cite{zhou2022rgb} partition subsets of the DSEC dataset\cite{gehrig2021dsec} for building object detection datasets. Tomy \emph{et al.}\cite{tomy2022fusing} annotates the training set of the DSEC dataset with three classes. However, this dataset contains considerable noise in the labels. Zhou \emph{et al.} \cite{zhou2022rgb} selects 16 sequences from the DSEC dataset to annotate bounding boxes for traffic motion objects. However, they annotate only 13,314 images without specifying object classes. Hence, a large-scale and high-quality RGB-Event traffic object detection dataset is still scarce. To address this problem, we manually annotate all sequences of the DSEC dataset and propose the DSEC-Det dataset, which has a larger scale, richer object classes, and higher annotation quality compared to the aforementioned dataset.

\subsection{Event-Based Object Detection}

\textbf{Event-based detection methods.} Benefiting from the outstanding attributes of event cameras, numerous event-based detection methods have been proposed \cite{perot2020learning,li2022asynchronous,wang2023dual,gehrig2023recurrent,wu2024leod}. RED \cite{perot2020learning} proposes a recurrent network architecture that uses ConvLSTM to extract spatiotemporal features from the event stream. ASTMNet\cite{li2022asynchronous} proposes a temporal attention convolutional module to learn event feature embedding from continuous event streams and a lightweight spatiotemporal memory module to extract temporal clues. DMANet\cite{wang2023dual} proposes a dual-memory aggregation network to leverage both long and short-term memory along event streams for effective spatial and temporal information aggregation. RVT\cite{gehrig2023recurrent} proposes a novel backbone for object detection that can reduce the inference time significantly while retaining similar performance to prior works. LEOD\cite{wu2024leod} proposes a self-training mechanism to enhance the utilization of unlabeled data. However, the lack of color and fine-grained texture information in event streams, which is crucial for achieving high detection performance, leads to decreased performance when compared to frame-based methods\cite{perot2020learning}.

\textbf{RGB and event fusion-based detection methods.} Detection by combining RGB and event streams is a reliable way to achieve high-performance object detection, which can be categorized into two strategies: late fusion and middle fusion. Late fusion methods include \cite{chen2018pseudo,jiang2019mixed,li2019event}. Chen \emph{et al.}\cite{chen2018pseudo} uses non-maximum suppression (NMS) to fuse the detection results of two modalities. Li \emph{et al.}\cite{li2019event} utilizes the Dempster-Shafer theory to fuse the detection results of the two modalities. While Jiang \emph{et al.}\cite{jiang2019mixed} proposes to fuse confidence maps of the two modalities. However, these late fusion methods lack intermodal interactions and fail to leverage the complementary nature effectively. Then, several middle fusion methods\cite{liu2021attention,cao2021fusion,tomy2022fusing,zhou2022rgb} are proposed to guide the fusion of two modalities on the feature level. Liu \emph{et al.}\cite{liu2021attention} computes a channel attention map to guide the learning of image features based on event features. Cao \emph{et al.}\cite{cao2021fusion} generates pixel-level attention maps based on features from two modalities and multiplies them with image features to obtain the fused features. Tomy \emph{et al.}\cite{tomy2022fusing} uses a simple concatenation operation to combine features from two modalities at different resolutions. Zhou \emph{et al.}\cite{zhou2022rgb} proposes a bidirectional fusion module to model multimodality features in the spatial and channel dimensions to form a shared representation. However, these methods cannot effectively perceive the information loss occurring in the images and adaptively compensate for it. To address this issue, we propose AFCM that perceives the global lightness distribution in images by calculating the global spatial correlation between the image and event modalities, and guides the interaction of the two modalities, allowing for a more comprehensive fusion of their respective strengths.

\subsection{Event Representation}
\textbf{Dense representation methods.} Since events are sparse impulse signals that cannot be directly processed using DNN methods, some researchers propose to convert the events into synchronous image-like tensors for processing. These methods can be roughly divided into image-based\cite{maqueda2018event,zhu2018ev,park2016performance}, surface-based\cite{lagorce2016hots,sironi2018hats,kim2021n}, and voxel-based\cite{zihao2018unsupervised}\cite{zhu2019unsupervised} methods and are widely applied to computer vision tasks such as image classification, action recognition, and object detection\cite{gallego2020event}. The image-based methods include Event Histogram\cite{maqueda2018event}, Event Image\cite{zhu2018ev} and Timestamp\cite{park2016performance}. While image-based approaches are straightforward to process, they tend to lose substantial temporal information. Surface-based methods, such as Time Surface\cite{lagorce2016hots}, HATS\cite{sironi2018hats}, and DiST\cite{kim2021n}, normalize timestamps to preserve temporal information of the event stream. Voxel-based methods, such as Voxel Grid\cite{zihao2018unsupervised}\cite{zhu2019unsupervised}, map original events to different time grids based on interpolation methods, to further preserve time dimension information. Nevertheless, they commonly opt for fixed time windows tailored to particular tasks and scenes, leading to low-quality structural information extraction during slow and fast-motion scenarios. Nikola \emph{et al.}\cite{zubic2023chaos} proposes the Gromov-Wasserstein Discrepancy (GWD) metric to measure the difference between raw events and their representations, and conducts hyperparameter search to find the optimal combination of the image-based methods. However, combining image-based methods can only partially alleviate this issue and cannot truly achieve robustness to complex motion patterns.

\textbf{Motion compensation methods.} These methods\cite{stoffregen2019event, gallego2019focus, xu2020robust, stoffregen2019event2, shiba2022event,shiba2022secrets, gallego2018unifying, hagenaars2021self, gu2021spatio, paredes2023taming,gehrig2021raft,Wu_2024_ICRA} can produce images of warped events (IWE) with sharp edges by estimating the motion vectors of events. 
They can be categorized into model-based methods\cite{gallego2018unifying,gu2021spatio,shiba2022secrets}, self-supervised learning methods\cite{hagenaars2021self,paredes2023taming}, and supervised learning methods\cite{gehrig2021raft,Wu_2024_ICRA}. Model-based methods\cite{gallego2018unifying,gu2021spatio,shiba2022secrets} simulate motion and optimize the model based on the principle of contrast maximization. Self-supervised learning methods\cite{hagenaars2021self,paredes2023taming} leverage the powerful modeling capabilities of deep neural networks to establish continuously-running stateful frameworks and learn to extract motion features from event data based on contrast-maximizing loss. Supervised learning methods\cite{gehrig2021raft,Wu_2024_ICRA} build networks inspired by frame-based literature and learn to estimate pixel motion from ground-truth data, outperforming model-based and self-supervised learning methods. While achieving impressive achievements, these methods still suffer from two issues: (1) Dependency on the assumption of constant illumination, where they assume that the scene illumination remains constant and events are solely generated by the motion. Autonomous driving scenarios with frequent changes in illumination (such as car headlights, streetlights, \emph{etc.}) can easily violate this assumption, and events caused by illumination changes can affect the performance of motion estimation\cite{shiba2022secrets}. (2) Dependency on the assumption of linear motion, where they assume that the motion of objects is linear. The motion estimation performance decreases when there are complex nonlinear motions or large pixel displacements\cite{paredes2023taming}.

\input{PDF/Figure/Fig_PiplineOfProposedSFNet}

\section{Proposed Methods}

This section first introduces the event imaging mechanism and output format in Section \ref{Event Camera}. Next, Section \ref{Overview} describes the overall architecture of the proposed method. Section \ref{RSGNet} introduces the details of the Reliable Structure Generation Network (RSGNet), uncovering its inner workings. Finally, Section \ref{AFCM} presents the design of the Adaptive Feature Complement Module (AFCM), 
shedding light on its purpose and functionality.

\subsection{Event Camera Introduction}\label{Event Camera}

The event camera outputs an event $(x_i,y_i,p_i,t_i)$ when the light intensity $\cal L$ changes at a pixel and exceeds the contrast threshold $c$. $(x_i,y_i)$ represents the spatial coordinates of pixel $i$, $t_i$ represents the timestamp when the event is triggered, and $p_i\in\{{-1,1}\}$ indicates the event polarity. The values 1 and -1 represent the increasing or decreasing change in intensity at the pixel, respectively. This process can be formulated as follows:
\begin{equation} 
{p_i} = \left\{ {\begin{array}{*{20}{l}}
{ + 1, \; {\rm{ if }} \; \log \left( {\frac{{{{\cal L}_t}\left( {{x_i},{y_i}} \right)}}{{{{\cal L}_{t - \Delta t}}\left( {{x_i},{y_i}} \right)}}} \right) > c,}\\
{ - 1, \; {\rm{ if }} \; \log \left( {\frac{{{{\cal L}_t}\left( {{x_i},{y_i}} \right)}}{{{{\cal L}_{t - \Delta t}}\left( {{x_i},{y_i}} \right)}}} \right) <  - c,}
\end{array}} \right.
\end{equation}
where ${{\cal L}_t}\left( {{x_i},{y_i}} \right)$ and ${{\cal L}_{t - \Delta t}}\left( {{x_i},{y_i}} \right)$ represent the intensity at $t$ and ${t - \Delta t}$, respectively.

\subsection{The Overview of the SFNet}\label{Overview}
In this section, we show the details of the Structure-aware Fusion Network (SFNet), which utilizes RGB and event modalities for traffic object detection in variable illumination conditions. As depicted in Fig. \ref{fig3}(a), the SFNet takes both an RGB image and the corresponding event stream as input. First, a novel Reliable Structure Generation Network (RSGNet) is designed to generate a Speed Invariant Frame (SIF) from the event stream, containing complete structures and sharp edges. Subsequently, the RGB image and SIF are separately transferred to two independent feature extractors to extract modality-specific features. In this paper, we empirically choose CSPDarkNet\cite{bochkovskiy2020yolov4} as the feature extractor. Further, to address information loss caused by non-uniform lighting, we propose an Adaptive Feature Completion Module (AFCM) to enable explicit interaction between the two modalities. As shown in Fig. \ref{fig3}(b-d), the AFCM comprises an Event Refine Module (ERM) that refines event features to suppress the influence of noise and a Lightness Distribution-aware Attention Module (LDAM) that adaptively fuses the event features into the image modality, enabling the modeling of illumination-robust feature representations. After that, the features with corresponding resolutions from both modalities are added and fed into FPN+PANet\cite{lin2017feature}\cite{liu2018path} for more comprehensive fusion at multiple semantic levels, fully leveraging their complementary characteristics. Finally, the fused features are fed into the YOLOX\cite{ge2021yolox} decoder to output class and bounding box information for each detected object.

\subsection{Reliable Structure Generation Network (RSGNet)}\label{RSGNet}

To achieve the fusion of images and events within a unified architecture, we opt to transform events into image-like tensors. Current image-like event representation methods often employ fixed time windows tailored to certain tasks and scenarios, resulting in sparse or blurry structural information for objects with varying speeds, which diminishes the expressiveness of the extracted features. To address this issue, we propose the RSGNet, which generates SIF with complete structures and sharp edges to accommodate diverse motion.

It’s important to note that the event output rate is directly proportional to the speed of motion and faster motion will output more events. Therefore, to guarantee information and structural integrity for all objects with varying speeds, we adopt a large time window for sampling. In this paper, we choose Timestamp \cite{park2016performance} to transfer event streams into frame $F\in{\mathbb{R}^{2 \times H \times W}}$. Timestamp partially retains time information while additionally capturing more details about slow-moving objects. Given the event stream $\left\{\left(x_{k}, y_{k}, t_{k}, p_{k}\right)\right\}_{k=1}^{N}$ within the time window $[0, T]$. $T$ is the time window duration which is set to 100 ms, and $N$ represents the number of events. The expression is as follows:
\begin{equation} 
{F_{x,y,p}} = \frac{{{{\max }_{x,y,p}}\{ {t_j}\} _{j = 1}^n}}{t_N},
\end{equation}
where $n$ is the number of events with polarity $p$ at $(x,y)$. In this way, complete structural information is extracted from the event stream. However, adopting a large time window inevitably results in elongated trajectories for fast-moving objects, leading to blurry edges in $F$.

To generate sharp edges from $F$, the discretized event polarity integration $E\in{\mathbb{R}^{B \times H \times W}}$ is used to assist the network. $E$ is computed following \cite{zhu2019unsupervised}, and $B$ is the number of time slices, which is set to 10. The expression is as follows:
\begin{equation} 
t_k^* = \frac{{B - 1}}{{{t_N} - {t_1}}}\left( {{t_k} - {t_1}} \right),
\end{equation}
\begin{equation} 
E_{x,y,t} = \sum\limits_k {{p_k}} \max \left( {0,1 - \left| {t - t_k^*} \right|} \right).
\end{equation}

This is based on the fact that the motion of objects can cause frequent brightness changes at the edges and trigger enormous events. Therefore, through the discretization of time and the accumulation of event polarities in each time slice, the discretized event polarity integration can reveal the continuous motion information at the edges of objects and expose their motion patterns, which can guide the sharp edges generation from $F$.

Finally, the $F$ and $E$ are concatenated and fed into an encoder-decoder network to generate the SIF. The learning process can be formulated as:
\begin{equation}
SIF = M(F,E;\theta ;S),
\end{equation}
where $M$ is the backbone of RSGNet, with U-Net\cite{ronneberger2015u} serving as $M$ for this paper. $\theta$ represents the learned parameters of $M$. The supervision signal $S$ is represented by the edge maps extracted from the RGB images using the Sobel operator, which offers simple processing and noise robustness.

In this paper, we utilize the local cross-correlation loss function to assess the sharpness of SIF. This loss function, compared to L1 and L2, possesses greater robustness in handling information disparities between event data and RGB. Thus it is particularly well-suited for measuring alignment between two distributions in non-aligned conditions. Meanwhile, in the process of aligning the distribution, it can utilize the local spatial smoothness inherent in the image modality to suppress noise in the event modality and facilitate subsequent fusion operations. The formula for this loss is as follows:

\begin{equation}
\begin{array}{*{20}{l}}
  {CC(SIF,S ) = } \\ 
  {\sum\limits_{{i}} {\frac{{{{\left( {\sum\limits_{{{q}} \in \Omega_{{i}}} {\left( {SIF\left( {{{{q}}}} \right) - \widehat {SIF}({{i}})} \right)} \left( {S\left( {{{{q}}}} \right) - \hat S ({{i}})} \right)} \right)}^2}}}{{\left( {\sum\limits_{{{q}} \in \Omega_{{i}}} {{{\left( {SIF\left( {{{{q}}}} \right) - \widehat {SIF}({{i}})} \right)}^2}} } \right)\left( {\sum\limits_{{{q}} \in \Omega_{{i}}} {{{\left( {S\left( {{{{q}}}} \right) - \hat S ({{i}})} \right)}^2}} } \right)}}} ,} 
\end{array}
\end{equation}

\begin{equation}
{\phi _{CC}}(SIF) =  -CC(SIF,S),
\end{equation}
where $\Omega$ is the local window size, $\widehat {SIF}({{i}})$ and $\hat S ({{i}})$ represent the local mean around the pixel $i$ in SIF and $S$, respectively.

In addition, we use TV loss to increase the contrast of SIF with the following equation:

\begin{equation}
{\phi _{{\text{TV}}}}(SIF) =  -(({\nabla_x SIF})^2 + ({\nabla_y SIF})^2),
\end{equation}
where ${\nabla_x}$ and ${\nabla_y}$ represent the horizontal and vertical gradient operations, respectively.

In summary, the total loss function of the RSGNet is:
\begin{equation}
{\phi _{{\text{loss}}}} = {\phi _{CC}}(SIF) + {\phi _{{\text{TV}}}}(SIF).
\end{equation}

Training strategy: RSGNet is devised to generate complete structures with sharp edges from the event stream, which requires complete structural information for supervision. However, due to the decreased contrast in improperly exposed RGB images, the extracted edges usually present incompleteness and noise. For these reasons, we select 4,933 images with normal exposure and extract their structural information to serve as the supervisory signal for RSGNet. The inherent high dynamic range nature of event cameras enables the generalization of RSGNet to different lighting conditions.

\input{PDF/Figure/Fig_OverviewOfOurDSEC-Det}

\input{PDF/Table/Tab_DatasetComparison}

\input{PDF/Figure/Fig_ProportionOfAnnotated}

\subsection{Adaptive Feature Completion Module (AFCM)}\label{AFCM}

In variable illumination traffic scenes, the structural properties of poorly exposed images are disrupted, and information loss occurs. Due to the influence of non-uniform lighting, the degree of information loss across images is non-uniform. To adaptively compensate for the non-uniform information loss, we propose an AFCM, which consists of an ERM to refine the event features and a LDAM to adaptively fuse features from the event modality into the image modality, modeling illumination-robust feature representations.

Considering that event cameras inevitably generate noise due to their sensitivity to junction leakage current and photocurrent, and the event noise will result in errors and uncertainties in the generated representations, especially in low-light scenarios\cite{wan2022s2n}\cite{ding2023mlb}. We design an ERM that leverages the local smoothness of the RGB features $f_R\in{\mathbb{R}^{C \times H \times W }}$ to eliminate the influence caused by event noise and refine the event features $f_E\in{\mathbb{R}^{C \times H \times W }}$. Specifically, a max pooling layer is applied along the channel dimension of $f_R$ to extract distinctive features. To suppress the noise in improperly exposed conditions, an average pooling layer is employed along the channel dimension to smooth the extracted information. Subsequently, the feature maps obtained from the max pooling and average pooling steps are concatenated and further processed to generate a smooth mask map. This map exhibits lower response values in the smooth areas of the RGB features, effectively suppressing noise caused by junction leakage current and photocurrent in the event modality. The formula for the ERM is as follows:

\begin{equation}
\widetilde{f_{E}} = {f_E} \otimes \sigma (Conv(Concat(APool(f_R);MPool(f_R)))),
\end{equation}
where APool is an average pooling layer, MPool is a max pooling layer, Concat indicates concatenation, Conv is a $7\times7$ convolution layer, $\sigma$ is the sigmoid function, $\otimes$ indicates element-wise multiplication, and $\widetilde{f_{E}}$ is the refined event features.

\input{PDF/Figure/Fig_DatasetAnnotationComparison}

Then, to achieve the adaptive fusion of the two modalities. We design a LDAM that computes the global spatial correlation $GSC$ between $f_R$ and $\widetilde{f_{E}}$ to perceive global lightness distribution in $f_R$, and adaptively extracts event features to compensate information loss. The LDAM is defined as:
\begin{equation}
{f_{R}'} = {\mathcal T}(Con{v_{1 \times 1}}({f_R})),
\end{equation}
\begin{equation}
{\widetilde{f_{E}}}_1 = {\mathcal T}(Con{v_{1 \times 1}}(\widetilde{f_{E}})),
\end{equation}
\begin{equation}
{\widetilde{f_{E}}}_2 = {\mathcal T}(Con{v_{1 \times 1}}(\widetilde{f_{E}})),
\end{equation}
\begin{equation}
{GSC} = Softmax ({f_{R}'} \cdot {\widetilde{f_{E}}}_1),
\end{equation}
\begin{equation}
{\widetilde{f_{E}}'} = \widetilde{f_{E}}_2 \cdot GSC,
\end{equation}
\begin{equation}
\widetilde{f_{R}} = {f_R} + {Con{v_{1 \times 1}}({\mathcal T}({\widetilde{f_{E}}'}))},
\end{equation}
where ${\mathcal T}$ represents the matrix reshape operation and $\cdot$ is matrix multiplication. The first three $1\times1$ convolution layers are used to reduce the number of channels of ${f_R}$ and ${\widetilde{f_{E}}}$. The event features ${\widetilde{f_{E}}'}$ required for RGB are generated through the multiplication of ${\widetilde{f_{E}}_2}$ by the global spatial correlation map. Following a channel dimension recovery using an additional $1\times1$ convolution, ${\widetilde{f_{E}}'}$ is combined with the original RGB features $f_R$ to yield the finalized RGB features $\widetilde{f_{R}}$.

\section{EXPERIMENTS}
This section first describes the details
of our proposed dataset. Then, we introduce the experimental settings. Subsequently, the quantitative and qualitative results are reported to demonstrate the effectiveness of our method. Finally, we conduct an ablation study on each module in our network.

\subsection{DSEC-Det Dataset}
Currently, some works\cite{tomy2022fusing}\cite{zhou2022rgb} propose RGB-Event object detection datasets based on the DSEC dataset\cite{gehrig2021dsec}. Zhou \emph{et al.}\cite{zhou2022rgb} selects 16 sequences from the DSEC dataset to introduce DSEC-MOD. However, this dataset is specifically designed for motion object detection tasks, solely providing labels for moving objects while omitting class information. Tomy \emph{et al.}\cite{tomy2022fusing} uses pretrained YOLOv5 to generate object detection annotations. However, this dataset annotates only 41 DSEC sequences with three classes and contains a significant amount of noise.

Motivated by the scarcity of event-based datasets with high-quality and large-scale annotations, we construct a dataset using all 53 sequences from the DSEC dataset, which offers rich and accurate object annotations, as shown in Fig. \ref{fig5}. The DSEC dataset is a large RGB-Event dataset in autonomous driving scenarios, collected under extremely challenging variable lighting conditions. Such complex lighting conditions present a significant challenge for frame-based cameras.

Specifically, we use homographic transformation based on camera matrices to align the viewpoint and resolution of the RGB and event cameras. Subsequently, we manually annotate all 53 sequences of the DSEC dataset, allocating 39 sequences for training and 14 for testing. For each sequence, we provide bounding box annotations for the following 8 classes, as shown in Fig. \ref{fig6}. The annotation file includes a list of labeled objects: each comprising the source image URL, the coordinates of the top-left of the bounding box, width, height, and a class label. Considering the small resolution of 640 × 480 of the event camera may impact the detection performance of small objects, we filter out bounding boxes with diagonals of less than 30 pixels following\cite{perot2020learning}.

Table \ref{table1} presents a comparison between the DSEC-Det dataset and existing event-based datasets. Notably, the DSEC dataset was recently expanded by the original team\cite{gehrig2021dsec}, incorporating additional sequences and utilizing a state-of-the-art object tracking algorithm for generating object detection annotations. However, the resulting labels also contain notable noise, as illustrated in Fig. \ref{fig4}. Such noise can significantly impact the model's performance. In contrast, our annotations produced manually by human annotators are more accurate and comprehensive.

\input{PDF/Table/Tab_ComparisonWithSotaObjectDet}

\input{PDF/Table/Tab_ComparisonWithSotaObjectDet_Exposure}

\input{PDF/Table/Tab_ComparisonWithSotaErAndMc}

\subsection{Experimental Settings}
\textbf{Datasets.} We evaluate our method on our proposed dataset DSEC-Det. In addition, to fairly compare with the state-of-the-art method RENet\cite{zhou2022rgb} that provides only their preprocessed results for the event modality without disclosing the processing code, we extract a subset from DSEC-Det called DSEC-Det-sub, consisting of sequences identical to those in DSEC-MOD. Furthermore, we divide the samples in DSEC-Det and DSEC-Det-sub into daytime and nighttime parts based on the exposure level. Specifically, the daytime part contains 9,968 frames from the DSEC-Det test set and 2,458 frames from the DSEC-Det-sub test set. The nighttime part contains 9,410 frames from the DSEC-Det test set and 366 frames from the DSEC-Det-sub test set. We perform the evaluation on both daytime and nighttime parts to demonstrate the effectiveness of our method under different lighting conditions.

From Fig. \ref{fig6}, we can observe the long tail properties, which present significant challenges. To evaluate the performance of our method under class-balanced and class-imbalanced conditions, we partition the dataset into two versions. The first version contains the primary classes, consisting of car and pedestrain, with a relatively balanced number between them. The second version contains all classes, with a significant class imbalance issue.

\textbf{Implementation Details.} We employ the Adam optimizer with a learning rate of 1e-4 and a batch size of 8 to train the RSGNet. Once trained, the weights of the RSGNet are held fixed. As for the detection network, we adopt YOLOX as the single modality baseline. The detection network is trained using the SGD optimizer with a learning rate of 5e-2 and a batch size of 6.

\textbf{Evaluation Metrics.} We adopt COCO metrics\cite{lin2014microsoft} to evaluate the performance scores, including mAP50 with an IOU threshold of 50\% and mAP50:95, which averages over IOUs between 50\% and 95\%. 

\subsection{Quantitative results}

To evaluate the superiority of our method, we compare our SFNet with SOTA (state-of-the-art) object detection methods, including event-based methods, frame-based methods, and RGB-Event fusion-based methods. Furthermore, to demonstrate the effectiveness of our event representation method, we conduct comparative experiments with SOTA event representation methods and motion compensation methods.

\textbf{Comparison with SOTA object detection methods.} We compare our method with five event-based methods: RVT\cite{gehrig2023recurrent}, DMANet\cite{wang2023dual}, Swinv2+YOLOv6\cite{zubic2023chaos}, YOLOX\cite{ge2021yolox}+Voxel\cite{zhu2019unsupervised} and YOLOX+SIF; six frame-based methods: Faster-RCNN\cite{ren2015faster}, RetinaNet\cite{lin2017focal}, CenterNet\cite{zhou2019objects}, YOLOv5\cite{Jocher_YOLOv5_by_Ultralytics_2020}, YOLOv7\cite{wang2023yolov7} and YOLOX; as well as two SOTA RGB-Event fusion-based object detection methods: FPN-fusion\cite{tomy2022fusing} and RENet\cite{zhou2022rgb}. These methods are trained on our dataset following their original training strategy to guarantee fairness.

\input{PDF/Figure/Fig_QualitativeComparisonWithRGB}

As depicted in Table \ref{table2}, our SFNet surpasses existing methods, encompassing both single modality and RGB-Event fusion-based methods. For example, in the class-imbalanced situation of the DSEC-Det dataset, our method outperforms the RGB baseline YOLOX by 3.9\% and FPN-fusion by 11.8\% in mAP50:95. Furthermore, the results of daytime and nighttime are shown in Table \ref{table22}. Poor exposure conditions at night result in lower image quality (reduced contrast and increased noise), which severely impacts the performance of frame-based methods. Owing to the high dynamic range capability of event cameras, our approach achieves superior performance in nighttime conditions, demonstrating the effectiveness of our method under varying lighting conditions. For example, our method outperforms the suboptimal method YOLOX by 7.8\% in mAP50 and 3.7\% in mAP50:95 in the class-imbalanced situation of the DSEC-Det dataset at night. Notably, despite incorporating the event modality, both FPN-fusion and RENet exhibit lower performance compared to frame-based methods. In contrast, our method achieves more robust object detection performance under variable illumination conditions through motion-robust event representation and effective cross-modality fusion for illumination-robust representations.

\textbf{Comparison with SOTA event representation methods and motion compensation methods.} To further demonstrate the superiority of our SIF, we compare it with five event representation methods: Timestamp\cite{park2016performance}, Time Surface\cite{lagorce2016hots}, DiST\cite{kim2021n}, Voxel\cite{zhu2019unsupervised} and ERGO-12\cite{zubic2023chaos}; seven motion compensation methods: (i) Model-based (MB): CMax\cite{gallego2018unifying}, ST-PPP\cite{gu2021spatio} and MCM\cite{shiba2022secrets}, (ii) Self-supervised learning (SSL): ConvGRU-EV-FlowNet\cite{hagenaars2021self} and Federico \emph{et al.} (configuration with the best performance) \cite{paredes2023taming}, (iii) Supervised learning (SL): E-RAFT\cite{gehrig2021raft} and IDNet (configuration with the best performance) \cite{Wu_2024_ICRA}. For motion compensation methods, the images of warped events (IWE) obtained after motion compensation serves as the input. We keep the detection network and fusion module the same and only change the input of event modality. The results are shown in Table \ref{table10}.

Among all the compared event representation methods, ERGO-12 achieves the best performance. However, our SIF outperforms ERGO-12 by producing sharper and more complete edge structures for objects with varying motion speeds while minimizing noise. For example, in the class-imbalanced situation of the DSEC-Det dataset, our SIF outperforms ERGO-12 by 2.3\% in mAP50.

Compared to model-based and self-supervised motion compensation methods, our SIF demonstrates superior performance. For the supervised methods E-RAFT and IDNet, it should be noted that they were trained on the DSEC-Flow dataset \cite{gehrig2021raft}, which overlaps with the test data of our DSEC-Det dataset, making the comparison unfair. Despite this, our SIF still outperforms them in most dataset settings. This success is attributable to RSGNet's ability to effectively manage various motion states of objects and its resilience to illumination changes. In contrast, diverse motion states and frequent illumination changes adversely affect these methods, resulting in decreased motion estimation accuracy and subsequent blurriness and distortion in the IWE, as shown in Fig. \ref{fig11}. Besides, we can observe that the best motion compensation method (\emph{i.e.} IDNet) does not achieve the best object detection performance, which indicates that the quality of IWE is not directly correlated with object detection performance.

\input{PDF/Figure/Fig_QualitativeComparisonWithEventRepresentation}

\input{PDF/Figure/Fig_QualitativeComparisonWithMotionCom}

\subsection{Qualitative results}
\textbf{Comparison with the RGB baseline.} We first present the comparison results between SFNet and RGB baseline YOLOX under different lighting conditions, as shown in Fig. \ref{fig7}. The first two rows show over-exposed scenes, and the subsequent two rows exhibit low-light scenes. By effectively leveraging the complementary nature of the two modalities and modeling illumination-robust feature representations, our method exhibits superior robustness compared to the RGB baseline in complex lighting scenes, enabling successful detection. Additionally, the final row shows a scene with motion blur. The high temporal resolution of event cameras enables precise capture of fast-moving car trajectories. Benefiting from the sharp structural information generated by our RSGNet, the category information of the car can be successfully identified.

\textbf{Comparison with SOTA event representation methods.} Then, we compare our SIF with Voxel \cite{zhu2019unsupervised}, Timestamp \cite{park2016performance}, Time Surface \cite{lagorce2016hots}, DiST \cite{kim2021n} and ERGO-12\cite{zubic2023chaos}, as shown in Fig. \ref{fig8}. The first two rows depict scenes with slowly moving objects, and the last row shows a scene with fast-moving objects. In the first two rows, the objects at the center exhibit a limited number of triggered events due to their slow motion relative to the camera and blending with the dark background. The existing event representation methods produce sparse contours that are prone to blending with background noise, resulting in detection failures. In the last row, existing methods produce blurry structures for the pedestrian that disrupt the structural information, resulting in detection failures. In contrast, our RSGNet can generate complete structures and sharp edges for both slow and fast moving objects, enabling successful detection.

\input{PDF/Figure/Fig_VisualizationOfAFCM}

\input{PDF/Table/Tab_PeformanceOfComponents}

\input{PDF/Table/Tab_PerformaceOfTimeWindow}

\input{PDF/Table/Tab_PerformaceOfSuperSignals}

\textbf{Comparison with SOTA motion compensation methods.}
Finally, we compare our SIF with CMax\cite{gallego2018unifying}, ST-PPP\cite{gu2021spatio}, MCM\cite{shiba2022secrets}, ConvGRU-EV-FlowNet\cite{hagenaars2021self}, Federico \emph{et al.}  \cite{paredes2023taming}, E-RAFT\cite{gehrig2021raft} and IDNet \cite{Wu_2024_ICRA}, as shown in Fig. \ref{fig11}. The performance of motion compensation methods can be significantly impacted by complex motion and frequent illumination changes, often leading to detection failures. Specifically, we can observe that: (a) In the first row, multiple fast-moving objects moving in the same direction are shown. A fast-moving car on the left overlaps its motion with a barrier, resulting in blurry edges; (b) The second row reveals the impact of fast-moving objects on slow-moving ones. A fast-moving car and a slow-moving car on the left create overlapping events, leading to an erroneous estimation of the slow-moving car's motion; (c) The third row presents the effect of nighttime noise on slow-moving objects. A slowly moving car on the left has its triggered events confused with background noise, resulting in failed motion estimation; (d) The fourth row illustrates the effect of local glare on objects. Events triggered by pedestrian movement overlap with those triggered by local glare, erroneously associating the former with the latter. Despite the challenges posed by complex motion and illumination changes, our RSGNet reliably generates sharp edges, enabling the detection framework to detect objects successfully.

\input{PDF/Table/Tab_ThePerformanceOfOurModelWithAFCM}

\input{PDF/Table/Tab_PerfromaceOfLoss}

\subsection{Ablation Study}
In this section, we present the ablation study results to demonstrate the effectiveness of each part of our SFNet.

\textbf{Contribution of our SFNet components.} We take YOLOX as the RGB baseline to perform a comprehensive ablation study and analyze the performance of each component in the SFNet. Table \ref{table3} shows the results from different combinations of modules in our method. We can observe that the model achieves higher performance with the assistance of the SIF. Combined with the AFCM, which consists of an ERM and a LDAM, our whole network significantly outperforms the baseline, validating the effectiveness of each proposed module under variable illumination traffic scenes.

The feature maps before and after applying the AFCM are illustrated in Fig. \ref{fig9} for three scenarios: (a) an over-exposure tunnel scene, (b) a nighttime scene with significant ground reflection interference, and (c) a nighttime scene with multiple objects. For the event modality, comparing the enhanced features ${\widetilde{f_{E}}}$ with the original ones ${f_{E}}$, we can observe that the ERM guides the event modality to re-integrate with the background and suppress noise. Similarly, for the image modality, comparing the enhanced features ${\widetilde{f_{R}}}$ with the original ones ${{f_{R}}}$, we can observe that the LDAM perceives regions of information loss caused by non-uniform lighting in the image modality and performs adaptive completion.

\textbf{Duration of the time window $T$.} To investigate the impact of different time window durations on detection performance, we conduct experiments comparing performance using various time windows. We use time windows of 50 ms, 100 ms, and 150 ms, corresponding to one, two, and three times the inter-frame interval of the image, respectively. As presented in Table \ref{table6}, the 100 ms time window delivers optimal performance, achieving a harmonious balance between information content and recovery complexity. The 50 ms short time window lacks sufficient data, hindering the extraction of complete structural details. Conversely, while the 150 ms long time window contains abundant information, its extended motion trajectories hinder effective blur removal. These factors impact the performance of the object detection network.

\textbf{Selection of the supervision signal $S$.} To assess the impact of different supervision signals on final detection performance, we conduct experimental comparisons of object detection results using the first-order differential operators Sobel and Roberts, as well as the second-order differential operator Laplace. The results present in Table \ref{table7} clearly demonstrate the significant superiority of the Sobel operator over the Roberts and Laplace operators. The performance gap can be attributed to the incompleteness of edge information extracted by the Roberts operator in the diagonal direction, and the noise introduced by the second-order differential operation of the Laplace operator. These factors can have an impact on the generation of clear and complete edge structures.

\input{PDF/Figure/Fig_VisualizationOfRSGNet}

\textbf{Loss functions in RSGNet.} To evaluate the superiority of the loss functions of our RSGNet, we compare the local cross-correlation loss with L1 and L2 losses. The results in Table \ref{table4} show that the local cross-correlation loss, which measures similarity at the distribution level, can better handle information disparities between event data and RGB than the pixel-wise L1 and L2 losses. The comparison of the first and last rows in Table \ref{table4} demonstrates that removing TV loss decreases performance, highlighting its importance for achieving high accuracy. Furthermore, Fig. \ref{fig10} visually demonstrates that adding TV loss significantly enhances the contrast of the SIF and the clarity of edges.

\textbf{AFCM placement.} Finally, we conduct an ablation study to analyze the relationship between detection performance and the AFCM placement. In this section, we use the SFNet without AFCM insertion as the baseline. Table \ref{table5} illustrates that placing the AFCM after the Conv and layer1 obtains the largest improvement compared with other placements. Naturally, the results agree with our motivation that completing the low-level features such as structural features adaptively in the image modality is beneficial.

\section{Conclusion}
In this paper, we propose a novel RGB-Event fusion architecture SFNet for traffic object detection in challenging illumination conditions, which consists of two key components: a RSGNet and an AFCM. The RSGNet can generate SIF with complete structures and sharp edges from the event stream to accommodate diverse motion. The AFCM can achieve adaptive fusion of the two modalities to address information loss caused by non-uniform lighting. Furthermore, we propose a large-scale DSEC-Det dataset for traffic object detection with rich and accurate annotations. The experimental results demonstrate that our method outperforms SOTA methods and boosts the robustness of traffic object detection in variable illumination conditions by leveraging the complementarities of the two modalities effectively. We focus on developing a multi-modal fusion framework to enhance the robustness of object detection in varying lighting conditions. We have yet to fully exploit the spatial sparsity and high temporal resolution of event cameras. Therefore, in the future, we will concentrate on the following two aspects: 1) We will explore a novel fusion framework that can achieve low-latency and high-frame-rate object detection by leveraging the spatial sparsity and high temporal resolution of event cameras. 2) We will construct a highly annotated RGB-Event dataset for intelligent transportation systems to further advance this field.

\section*{Acknowledgment}
This research is supported by the National Key Research and Development Program of China [2023YFC3081700], the National Natural Science Foundation of China (General Program) [No. 52172302], the Two-chain Integration Key Special Project of Shaanxi Provincial Department of Science and Technology - Enterprise-Institute Joint Key Special Project [2023-LL-QY-24], and Shannxi Province Traffic Science and Technology Program [21- 02X].

\bibliographystyle{IEEEtran}
\bibliography{reff} 

\vspace{-40pt}
\begin{IEEEbiography}[{\includegraphics[width=1in,height=1.25in,clip,keepaspectratio]{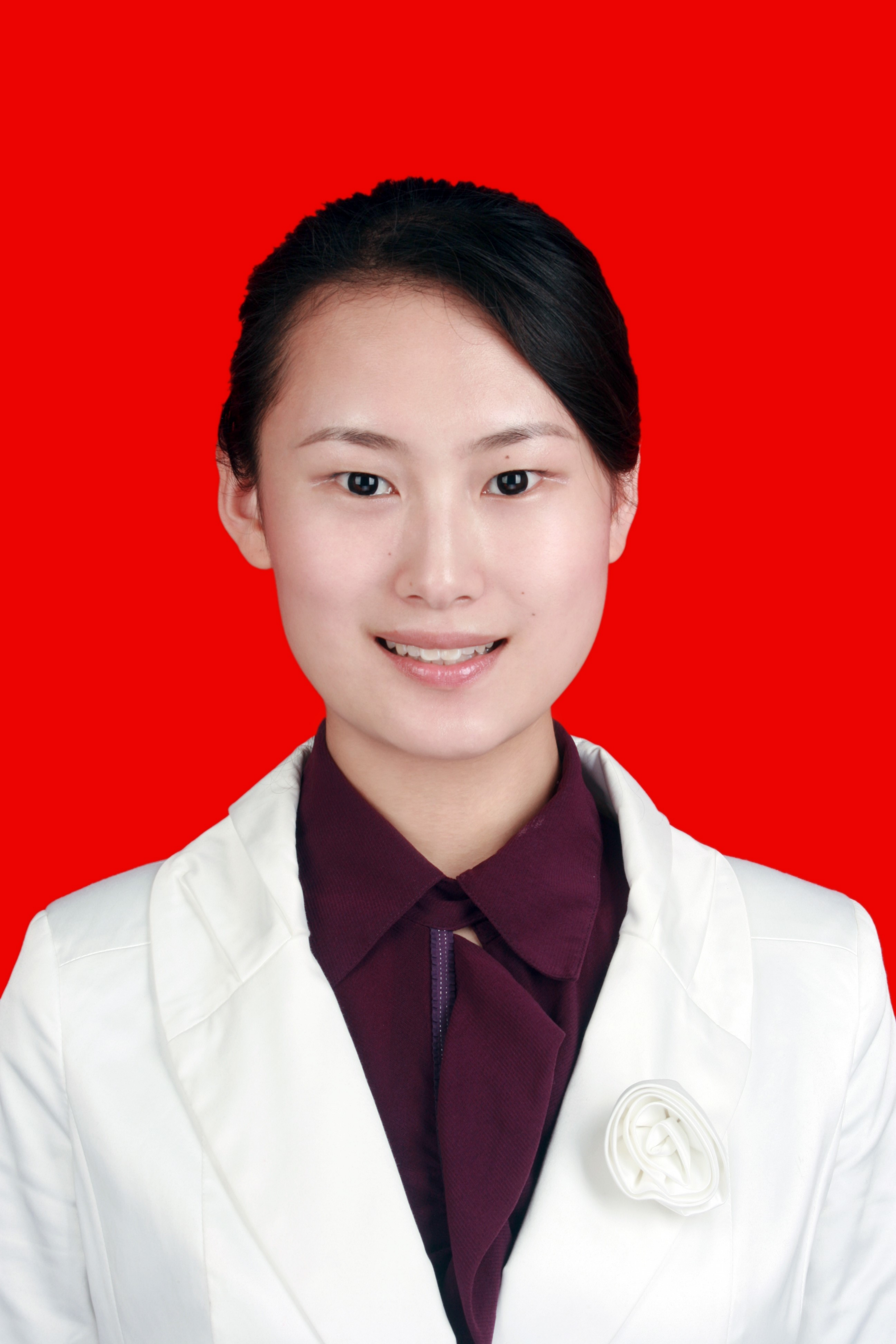}}]{Zhanwen Liu}
(Member, IEEE) received the B.S. degree from Northwestern Polytechnical University, Xi'an, China, in 2006, the M.S. and the Ph.D. degrees in Traffic Information Engineering and Control from Chang'an University, Xi'an, China, in 2009 and 2014 respectively. She is currently a professor with School of Information Engineering, Chang'an University, Xi'an, China. Her research interests include vision perception, autonomous vehicles, deep learning and intelligent transportation systems.
\end{IEEEbiography}

\vspace{-30pt}

\begin{IEEEbiography}
[{\includegraphics[width=1in,height=1.25in,clip,keepaspectratio]{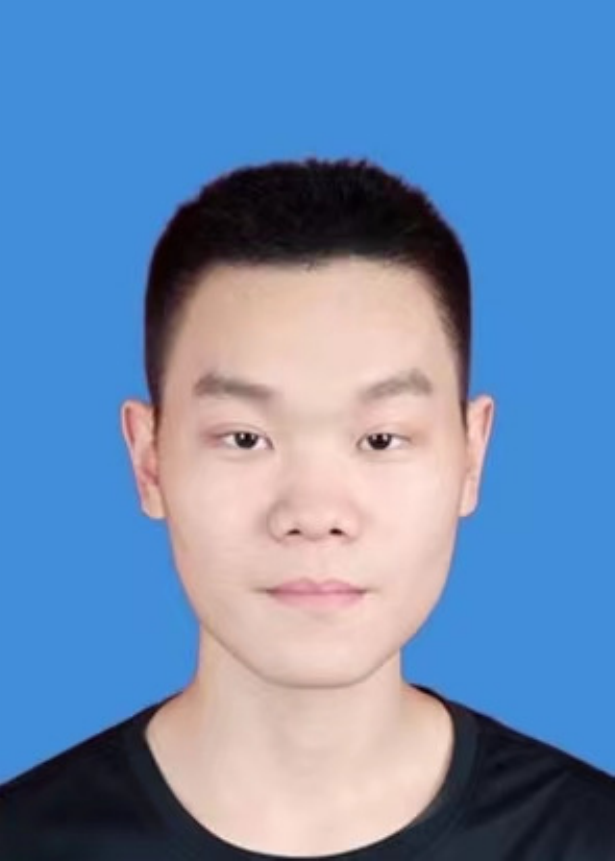}}]{Nan Yang}
received the B.S. degree from Chang’an University in Xi’an, China, in 2022, where he is currently working toward the Ph.D. degree in Traffic Information Engineering and Control. His current research interests include object detection and multiple-object tracking based on event cameras, and their applications in intelligent vehicle and road infrastructure perception.
\end{IEEEbiography}

\vspace{-30pt}

\begin{IEEEbiography}
[{\includegraphics[width=1in,height=1.25in,clip,keepaspectratio]{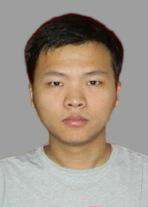}}]{Yang Wang}
(Member, IEEE) received the B.S. degree from Chang’an University, Xi’an, China, in 2016, the Ph.D. degree in control science and engineering from the University of Science and Technology of China, in 2021. He is currently an Associate professor with the School of Information Engineering, Chang’an University, Xi’an, China. His central research interests focus on computer vision and multimedia processing.
\end{IEEEbiography}

\vspace{-30pt}
\begin{IEEEbiography}
[{\includegraphics[width=1in,height=1.25in,clip,keepaspectratio]{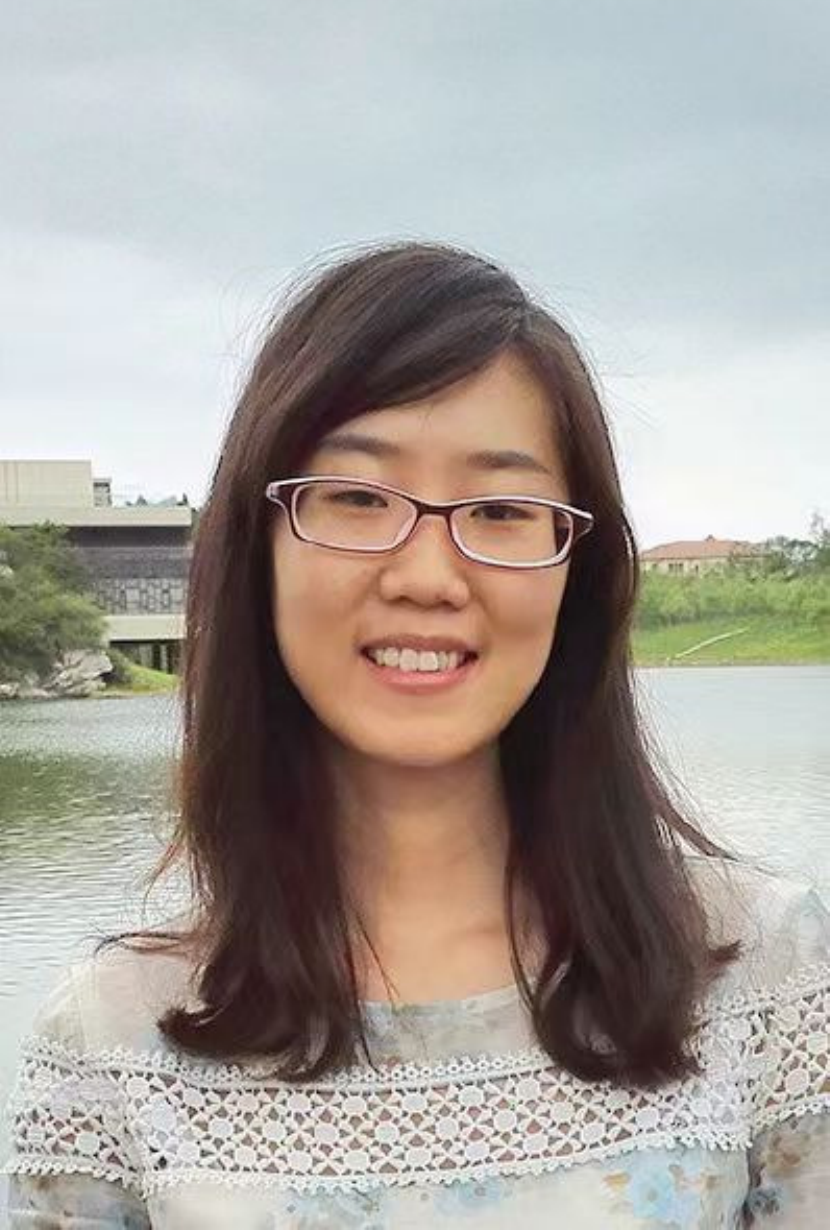}}]{Yuke Li}
received the B.E. degree in Communication Engineering from Xidian University, Xi’an, China, in 2014, the Ph.D. degree in Control Theory and Control Engineering at the State Key Laboratory of Management and Control for Complex Systems, Institute of Automation, Chinese Academy of Sciences, Beijing, China, in 2019. From November 2016 to November 2017, she was a Visiting Scholar with Colorado State University, Fort Collins, CO, USA. From October 2021 to October 2023, she was a Postdoctoral Researcher at Huawei Technologies Co., Ltd. Now she is a Senior Engineer at Waytous Co. Ltd., Beijing, China. Her research interests include intelligent transportation systems, parallel intelligence, signal processing, and joint radar and communication. She received the Best Paper Award from the IEEE International Conference on Intelligent Transportation Systems in 2014, Beijing Outstanding Graduates Award in 2019, and Huawei Outstanding Postdoctoral Fellow in 2023.
\end{IEEEbiography}

\vspace{-30pt}
\begin{IEEEbiography}
[{\includegraphics[width=1in,height=1.25in,clip,keepaspectratio]{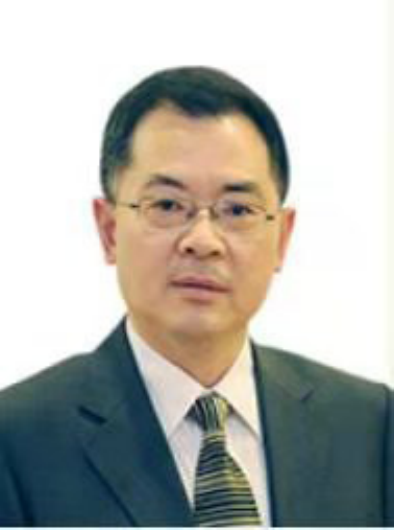}}]{Xiangmo Zhao}
(Member, IEEE) received the Ph.D. degree from Chang’an University, Xi’an, China. He is currently a Professor with the School of Information Engineering, Chang’an University. He is currently the Vice President of the Joint Laboratory for Connected Vehicles, Ministry of Education-China Mobile Communications Corporation, and the Shaanxi Road Traffic Intelligent Detection and Equipment Engineering Technology Research Centre, and also the Leader of the National Key Subjects-Traffic Information Engineering and Control, Chang’an University. His current research interests include connected vehicles, automated vehicles, intelligent transportation systems, and computer science. He is the Director of the Information Professional Committee; a member of the Advisory Expert Group of the China Transportation Association, the National Motor Vehicle Operation Safety Testing Equipment Standardization Committee, and the Leading Group of the National Traffic Computer Application Network; the Vice Chairperson of the Institute of Highway Association on Computer Professional Committee; and the Deputy Director of the Institute of Computer in Shaanxi Province.
\end{IEEEbiography}

\begin{IEEEbiography}
[{\includegraphics[width=1in,height=1.25in,clip,keepaspectratio]{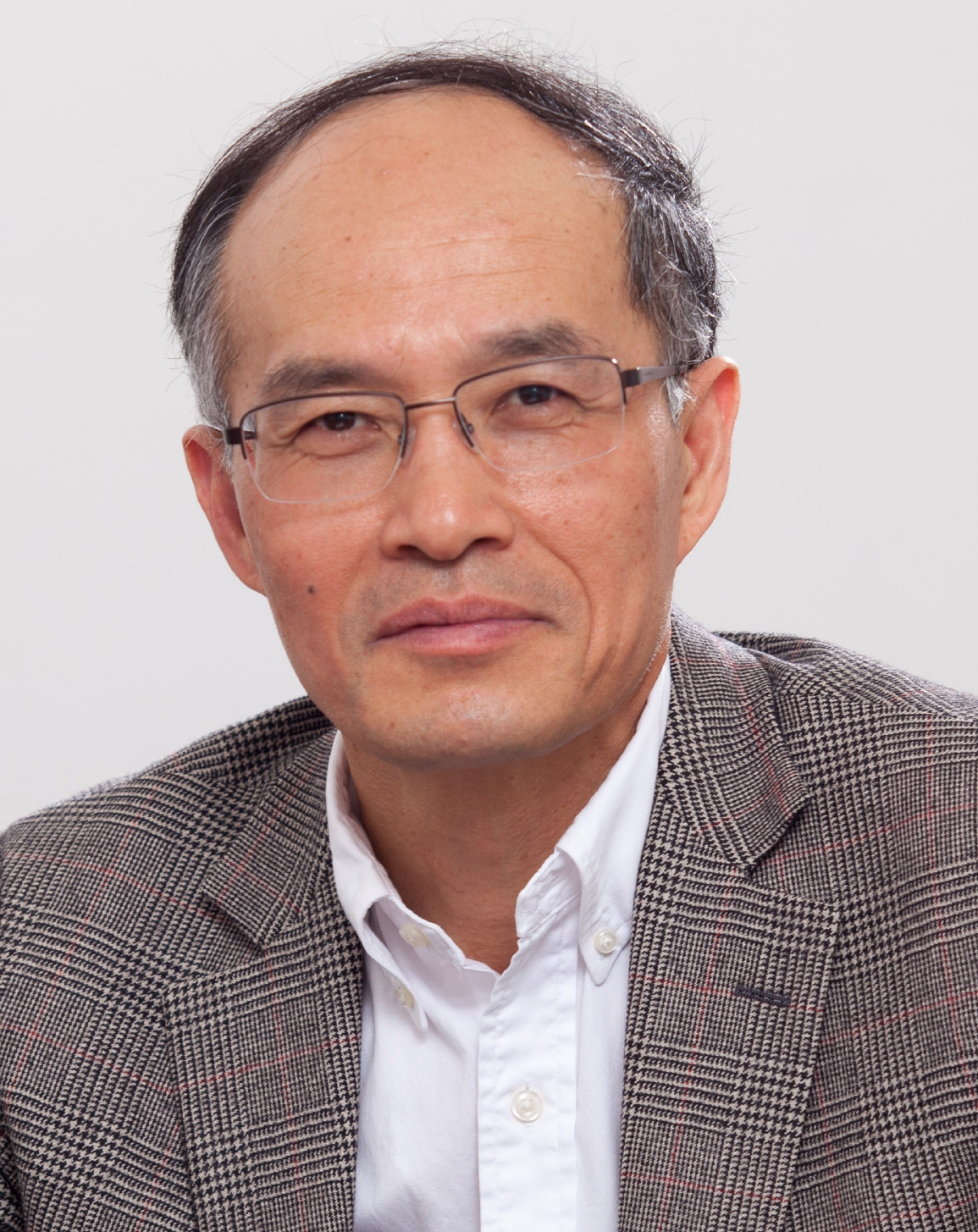}}]{Fei-Yue Wang}
(Fellow, IEEE) received the Ph.D. degree in computer and systems engineering from the Rensselaer Polytechnic Institute, Troy, NY, USA, in 1990. 

He joined The University of Arizona in 1990 and became a Professor and the Director of the Robotics and Automation Laboratory and the Program in Advanced Research for Complex Systems. In 1999, he founded the Intelligent Control and Systems Engineering Center, Institute of Automation, Chinese Academy of Sciences (CAS), Beijing, China, under the support of the Outstanding Chinese Talents Program from the State Planning Council. In 2002, he was appointed as the Director of the Key Laboratory for Complex Systems and Intelligence Science, CAS, where he was the Vice President of the Institute of Automation, in 2006. He found CAS Center for Social Computing and Parallel Management in 2008, and became the State Specially Appointed Expert and the Founding Director of the State Key Laboratory for Management and Control of Complex Systems in 2011. His current research focuses on methods and applications for parallel intelligence, social computing, DeSci, and knowledge automation.

He is a fellow of INCOSE, IFAC, ASME, and AAAS. In 2007, he received the National Prize in Natural Sciences of China and numerous best papers awards from IEEE TRANSACTIONS. He became an Outstanding Scientist of ACM for his work in intelligent control and social computing. He received the IEEE ITS Outstanding Application and Research Awards in 2009, 2011, and 2015; and the IEEE SMC Norbert Wiener Award in 2014. Since 1997, he has been serving as the general chair or the program chair for over 30 IEEE, INFORMS, IFAC, ACM, and ASME conferences. He was the President of the IEEE ITS Society, from 2005 to 2007; the IEEE Council of RFID, from 2019 to 2021; the Chinese Association for Science and Technology, USA, in 2005; the American Zhu Kezhen Education Foundation, from 2007 to 2008. He was the Vice President of the ACM China Council, from 2010 to 2011; and the IEEE Systems, Man, and Cybernetics Society, from 2019 to 2021. He was the Vice President and the Secretary General of the Chinese Association of Automation, from 2008 to 2018. He was the Founding Editor-in-Chief (EiC) of the International Journal of Intelligent Control and Systems, from 1995 to 2000; IEEE Intelligent Transportation Systems Magazine, from 2006 to 2007; IEEE/CAA JOURNAL OF AUTOMATICA SINICA, from 2014 to 2017; China’s Journal of Command and Control, from 2015 to 2021; and China’s Journal of Intelligent Science and Technology, from 2019 to 2021. He was the EiC of the IEEE INTELLIGENT SYSTEMS, from 2009 to 2012, IEEE TRANSACTIONS ON INTELLIGENT TRANSPORTATION SYSTEMS, from 2009 to 2016, and IEEE TRANSACTIONS ON COMPUTATIONAL SOCIAL SYSTEMS, from 2017 to 2020. Currently, he is the President of the CAA’s Supervision Council, and the EiC of IEEE TRANSACTION ON INTELLIGENT VEHICLES. He became the IFAC Pavel J. Nowacki Distinguished Lecturer in 2021.
\end{IEEEbiography}

\end{document}

%% file: PDF/Figure/Fig_FeatureAfterFusion.tex
\begin{figure}[tbp]
\centering
\includegraphics[scale=0.177]{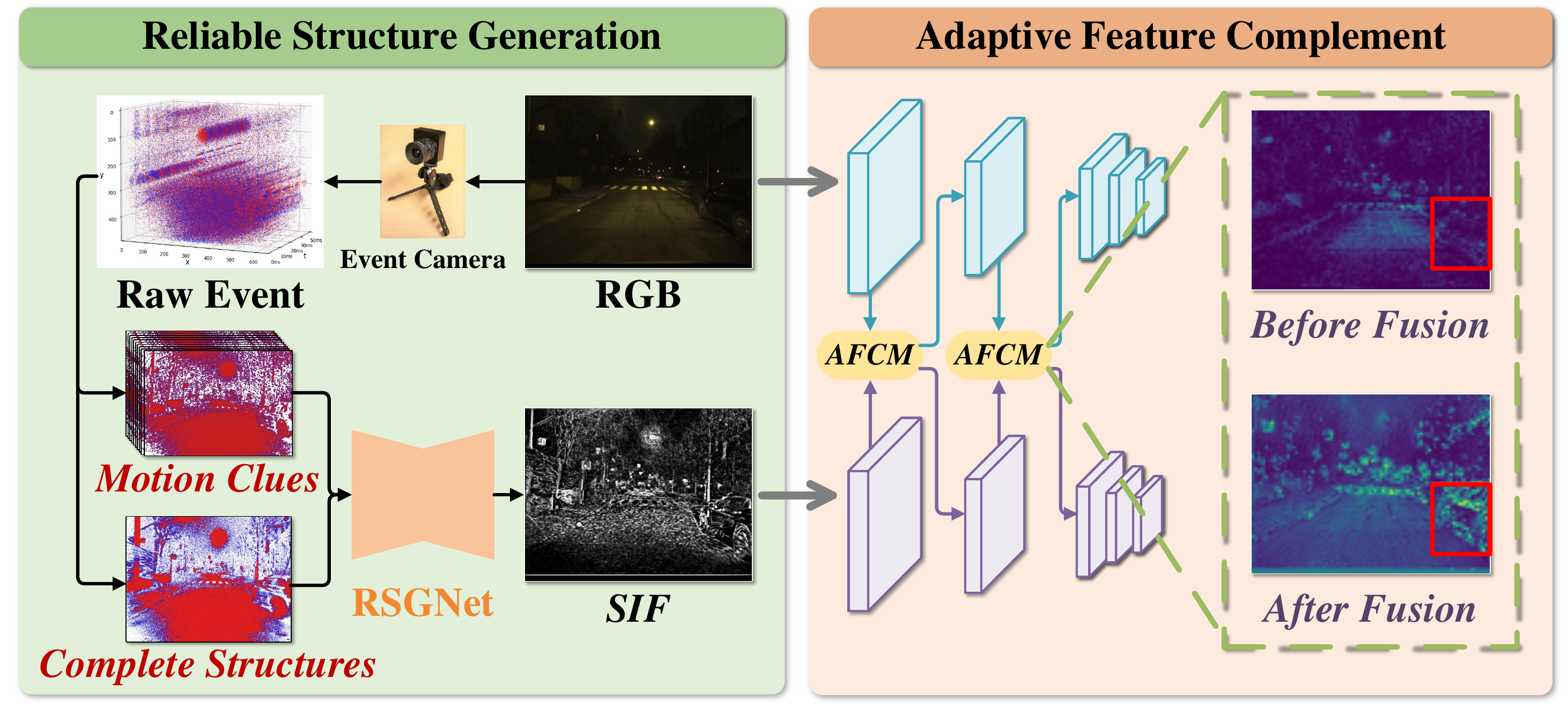}
\caption{Our SFNet comprises two cascaded steps: reliable structure generation and adaptive feature complement. Left: Given an RGB image and corresponding event stream, high-quality structural information is generated from the event stream. Right: The event feature is adaptively extracted to compensate for information loss in the image modality. The image feature, before complementation, lacks discriminative characteristics due to information loss and contrast reduction across the image. Notable, after the complementation, the feature becomes more discriminative.}
\label{fig1}
\end{figure}

%% file: PDF/Figure/Fig_FixedTimeWindows.tex
\begin{figure}[tbp]
\centering
\includegraphics[scale=0.27]{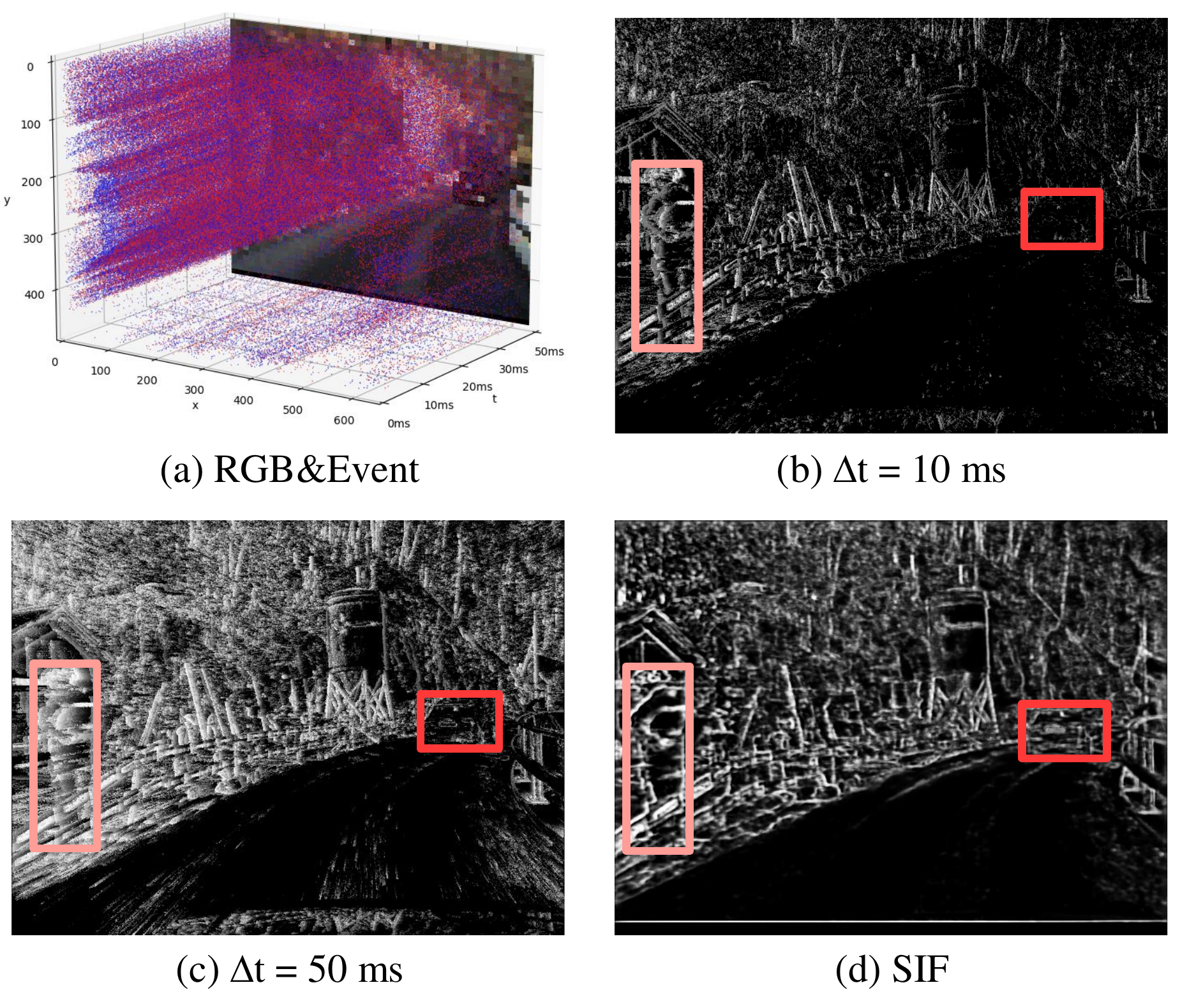}
\caption{Fixed time windows for event representation. The pedestrian nearby moves at a faster speed compared to the car in the distance. (b) displays a short time window of $\Delta t = 10 \; ms$, resulting in sparse structures for the car. Conversely, (c) shows a long time window of $\Delta t = 50 \; ms$, resulting in blurry structures for the pedestrian. Notably, our Speed Invariant Frame in (d) shows the complete structures and sharp edges for both pedestrian and car. We utilize \textcolor[RGB]{255,56,56}{\textbf{red}} and \textcolor[RGB]{255,175,252}{\textbf{pink}} boxes to mark the car and pedestrian, respectively.}
\label{fig2}
\end{figure}

%% file: PDF/Figure/Fig_PiplineOfProposedSFNet.tex
\begin{figure*}[htbp]
\centering
\includegraphics[scale=0.155]{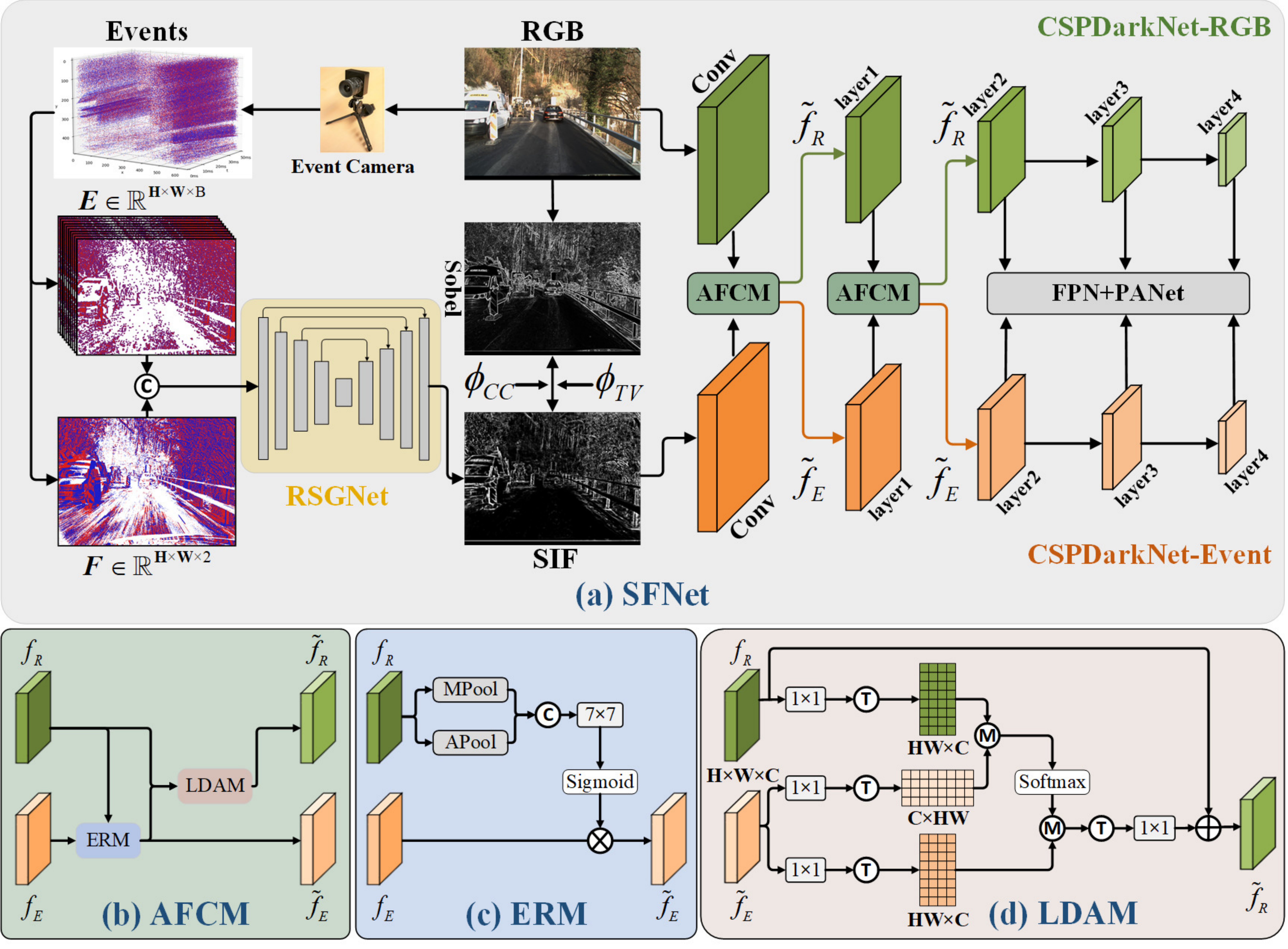}
\centering
\caption{The pipeline of the proposed Structure-aware Fusion Network (SFNet). The event frame $F$ and event polarity integration $E$ generated from the event stream are concatenated and fed into the Reliable Structure Generation Network (RSGNet see Section \ref{RSGNet}) to generate the Speed Invariant Frame (SIF). Then the RGB image and SIF are respectively input into two independent CSPDarkNets to extract modality-specific features. The Adaptive Feature Completion Module (AFCM see Section \ref{AFCM}), which consists of an Event Refine Module (ERM) and a Lightness Distribution-aware Attention Module (LDAM) is inserted after the Conv and layer1 stages to perform the fusion of the two modalities. FPN+PANet fuses the features at different resolutions further. Finally, the decoder outputs each detected object's class and bounding box information.
}
\label{fig3}
\end{figure*}

%% file: PDF/Figure/Fig_OverviewOfOurDSEC-Det.tex
\begin{figure*}[htbp]
\centering
\includegraphics[scale=0.235]{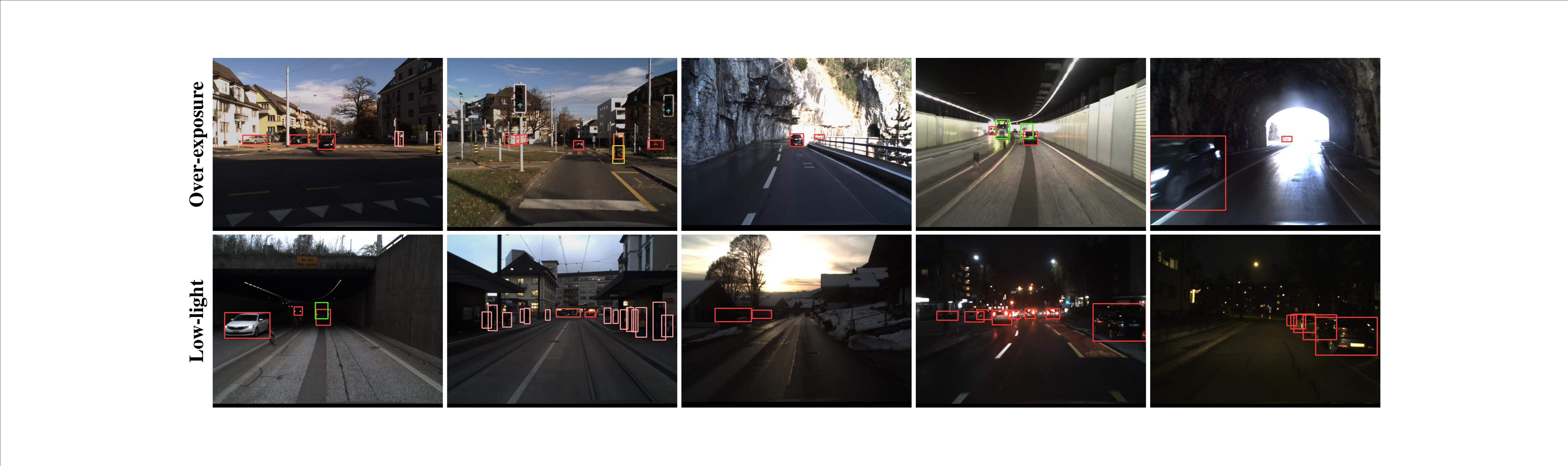}
\caption{Overview of our DSEC-Det dataset, which contains extremely challenging variable lighting conditions and rich object annotations.}
\label{fig5}
\end{figure*}

%% file: PDF/Table/Tab_DatasetComparison.tex
\begin{table*}[htbp]
\centering
  \caption{Comparison Of Different Event-based Object Detection Datasets In Driving Scenarios. “Automated” represents the annotations generated by a pretrained model. “Manual” represents the annotations produced manually by human annotators, which is more accurate.}
  \label{tab:commands}
  \footnotesize
  \renewcommand\arraystretch{1.3}
  \begin{tabular} {c| p{1cm}<{\centering} p{1cm}<{\centering} p{2cm}<{\centering} p{2.5cm}<{\centering} p{2.5cm}<{\centering} p{1cm}<{\centering} p{1cm}<{\centering}}
    \Xhline{1.5pt}
    Dataset & Intensity & Class & Resolution & Frames & Labeling method & Real & Public\\
    \hline
    Gen1\cite{de2020large} & {\ding{55}} & 2 & 304×240 & 16,984,800(39.32h) & Manual & {\checkmark} & {\checkmark} \\
    1 Mpx\cite{perot2020learning} & {\ding{55}} & 7 & 1280×720 & 3,164,400(14.65h) & Automated & {\checkmark} & {\checkmark} \\
    \hline
    EventKITTI\cite{liang2022global} & RGB & 8 & 1333×401  & 7,481& Manual & {\ding{55}} & {\ding{55}} \\
    DSEC-MOD\cite{zhou2022rgb} & RGB & 1 & 640×480  & 13,314& Manual & {\checkmark} & {\checkmark} \\
    Tomy \emph{et al.}\cite{tomy2022fusing} & RGB & 3 & 640×480 & 44,776& Automated & {\checkmark} & {\checkmark}\\
    DSEC-Detection\cite{gehrig2021dsec} & RGB & 8 & 640×480 & 70,379& Automated & {\checkmark} & {\checkmark}\\
    Ours & RGB & 8 &640×480 & 63,931& Manual
 &  {\checkmark} & {\checkmark}\\
    \Xhline{1.5pt}
  \end{tabular}
  \label{table1}
\end{table*}

%% file: PDF/Figure/Fig_ProportionOfAnnotated.tex
\begin{figure}[tbp]
\centering
\includegraphics[scale=0.20]{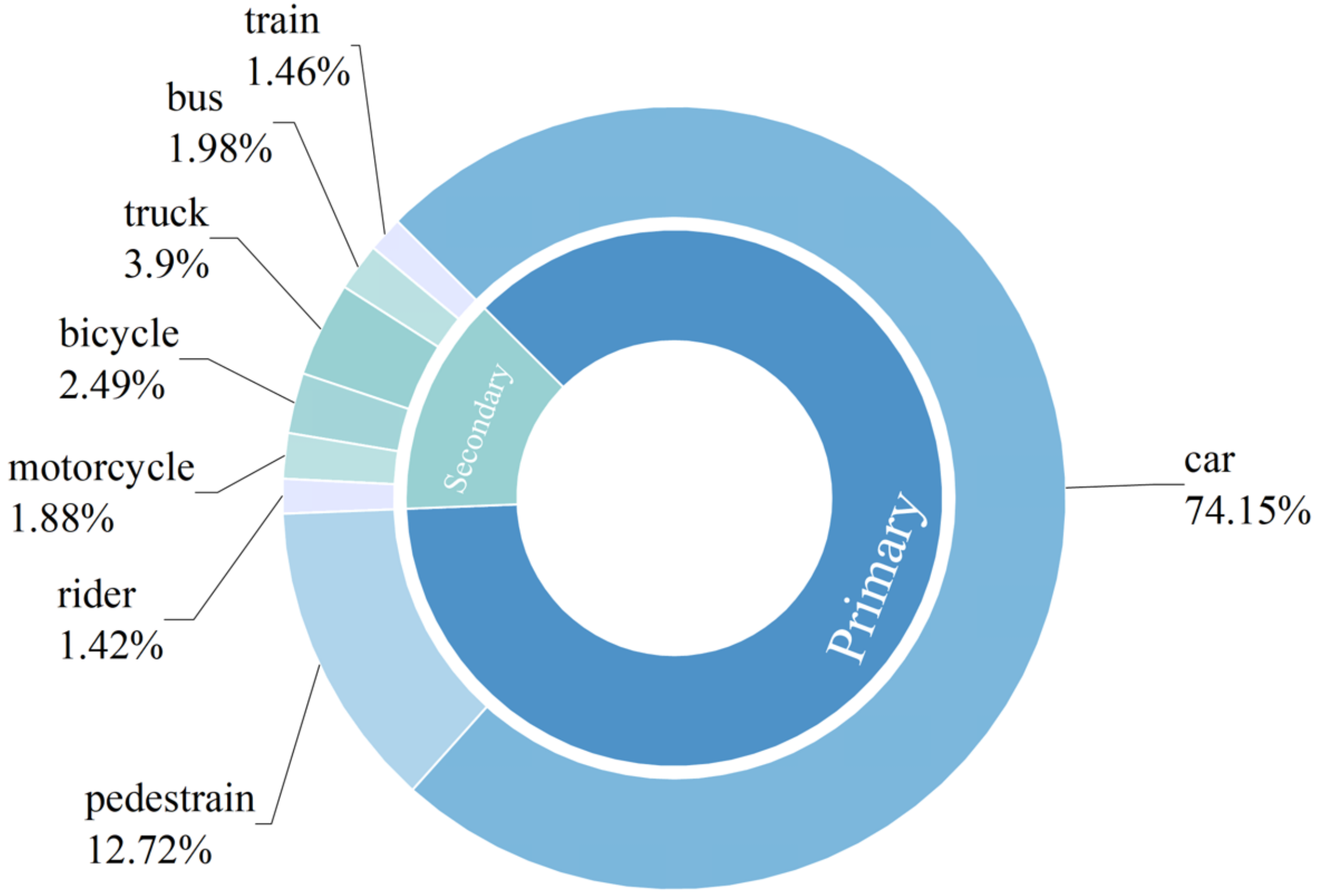}
\caption{Proportion of annotated bounding boxes for the DSEC-Det dataset.}
\label{fig6}
\end{figure}

%% file: PDF/Figure/Fig_DatasetAnnotationComparison.tex
\begin{figure*}[tbp]
\centering
\includegraphics[scale=0.355]{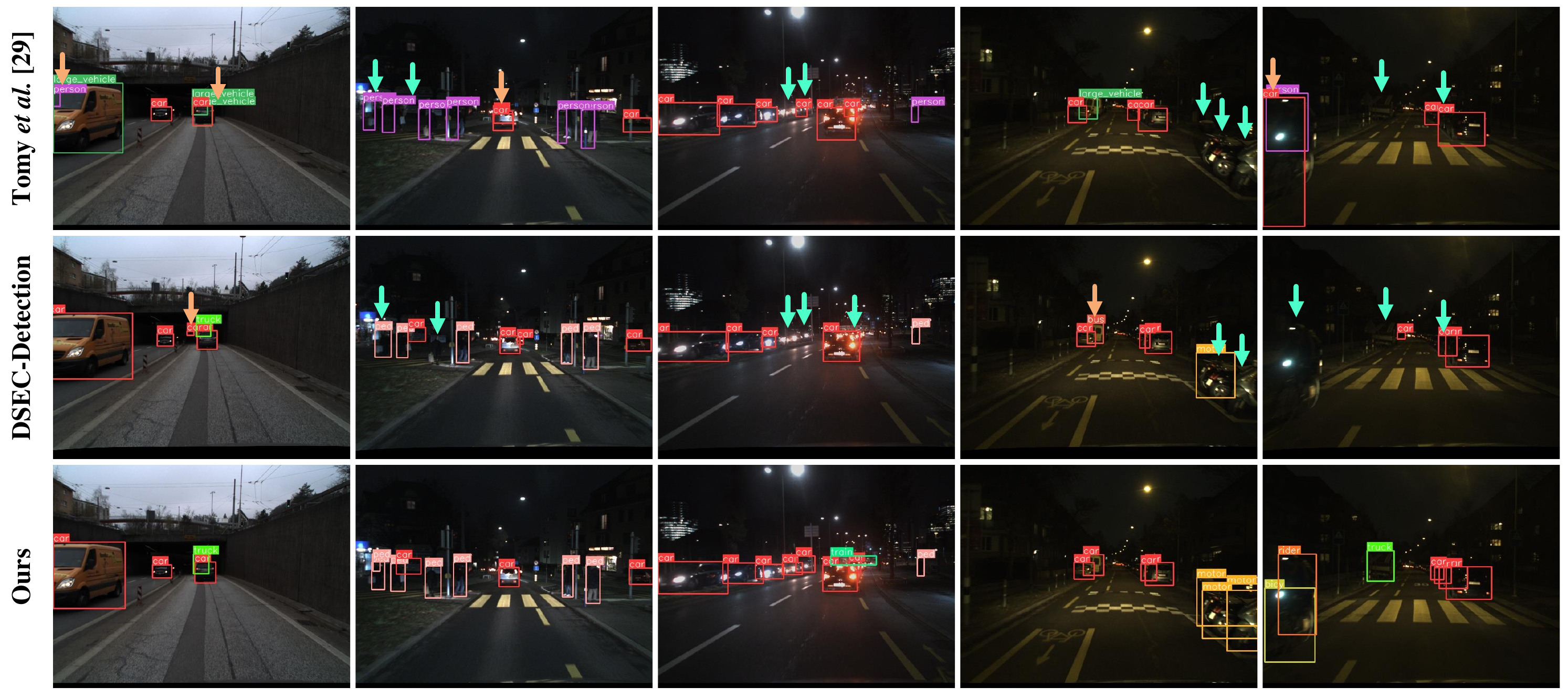}
\caption{Comparison with Tomy \emph{et al.} \cite{tomy2022fusing} and DSEC-Detection\cite{gehrig2021dsec}, whose annotations are generated by pretrained models. Both of them have cases of incorrect labels, especially in low-light conditions, and there is an offset between objects and bounding boxes in \cite{tomy2022fusing}. In contrast, our annotations produced manually by human annotators demonstrate higher accuracy and completeness. We utilize \textcolor[RGB]{11,255,200}{\textbf{green}} and \textcolor[RGB]{248,171,114}{\textbf{orange}} arrows to mark the missed and false detections.}
\label{fig4}
\end{figure*}

%% file: PDF/Table/Tab_ComparisonWithSotaObjectDet.tex
\begin{table*}[tbp]
\belowrulesep=0pt
\aboverulesep=0pt
\centering
  \setlength{\tabcolsep}{1.1mm}
  \caption{Performance comparison with SOTA methods. The best and the second-best performances are
 marked in {\color{red}{\textbf{red bold}}} and {\color{blue}{\textbf{blue bold}}}, respectively.}
  \label{tab:commands}
  \renewcommand\arraystretch{1.7} 
  \scalebox{0.8}{
  \begin{tabular} {cccc|cc|cc|cc|cc}
    \Xhline{1.5pt}
    {\multirow{3}*{{\large Model Type}}} & {\multirow{3}*{{\large Method}}} & {\multirow{3}*{{\large Input}}} & {\multirow{3}*{{\large Pub. \& Year}}} & \multicolumn{4}{c|}{{\large DSEC-Det}} & \multicolumn{4}{c}{{\large DSEC-Det-sub}} \\ 
    & & & & \multicolumn{2}{c}{{\normalsize Class-balanced}} &  \multicolumn{2}{c|}{{\normalsize Class-imbalanced}} & \multicolumn{2}{c}{{\normalsize Class-balanced}} &  \multicolumn{2}{c}{{\normalsize Class-imbalanced}}\\
    \hhline{~~~~--------}
    & & & & { mAP50} & { mAP50:95} & { mAP50}& { mAP50:95} & { mAP50} & { mAP50:95} & { mAP50}& { mAP50:95}\\
    \hline
    
    {\multirow{5}*{{\large \rotatebox{90}{Event}}}} & {\large RVT\cite{gehrig2023recurrent}} & {\large Voxel\cite{zhu2019unsupervised}} & {\normalsize {CVPR'2023}} & {\large {51.1}} & {\large 26.6} & {\large {25.1}} & {\large 12.9} & {\large {50.9}} & {\large 26.0} & {\large {20.3}} & {\large {10.7}}\\ 
    
    & {\large DMANet\cite{wang2023dual}} & {\large EventPillars\cite{wang2023dual}} & {\normalsize {AAAI'2022}} & {\large 29.7} & {\large 18.1} & {\large 13.5} & {\large 6.8} & {\large 36.5} & {\large 19.2} & {\large 13.8} & {\large 6.8}\\
    
    & {\normalsize Swinv2+YOLOv6\cite{zubic2023chaos}} & {\large ERGO-12\cite{zubic2023chaos}} & {\normalsize {ICCV'2023}} & {\large 2.2} & {\large 1.0} & {\large 6.8} & {\large 3.2} & {\large 9.6} & {\large 4.2} & {\large 12.8} & {\large 6.5}\\
    
    & {\large YOLOX\cite{ge2021yolox}} & {\large Voxel\cite{zhu2019unsupervised}} & {\normalsize {arXiv'2021}} & {\large 49.2} & {\large 25.3} & {\large 23.9} & {\large 11.7} &  {\large 46.1} & {\large 22.9} & {\large 16.8} & {\large 8.4}\\
    
    & {\large YOLOX\cite{ge2021yolox}} & {\large SIF} & {\normalsize {arXiv'2021}} & {\large 55.7} & {\large 29.9} & {\large 23.8} & {\large 12.6} &  {\large 52.6} & {\large 30.7} & {\large 20.4} & {\large 11.0}\\
   
    \hline
    \hline
    {\multirow{5}*{{\large \rotatebox{90}{RGB}}}} & {\large Faster-RCNN\cite{ren2015faster}} & {\large RGB}  & {\normalsize {NeurIPS'2015}}& {\large {\color{blue} \textbf{76.7}}} & {\large 44.5} & {\large {\color{blue} \textbf{45.7}}} & {\large 24.9} & {\large {\color{blue} \textbf{73.7}}} & {\large \color{blue} \textbf{44.6}} & {\color{red} \textbf{\large {38.0}}} & {\large {21.7}}\\ 
    
    & {\large RetinaNet\cite{lin2017focal}} & {\large RGB}  & {\normalsize {ICCV'2017}}& {\large 61.3} & {\large 33.9} & {\large 36.0} & {\large 20.2} & {\large 65.2} & {\large 39.5} & {\large 37.2} & {\large 20.3}\\ 
    
    & {\large CenterNet\cite{zhou2019objects}} & {\large RGB}  & {\normalsize {arXiv'2019}}& {\large 73.4} & {\large 43.6} & {\large 44.4} & {\large 23.9} & {\large 72.7} & {\large \color{blue} \textbf{44.6}} & {\large 36.1} & {\large 19.0}\\
    
    & {\large YOLOv5\cite{Jocher_YOLOv5_by_Ultralytics_2020}} & {\large RGB}  & {\normalsize {2020}}& {\large 75.5} & {\large {48.0}} & {\large 42.2} & {\large {26.4}} & {\large 70.3} & {\large {44.4}} & {\color{blue} \large{\textbf{37.5}}} & {\large 21.8}\\ 
    
    & {\large YOLOv7\cite{wang2023yolov7}} & {\large RGB}  & {\normalsize {CVPR'2023}}& {\large 76.2} & {\large {46.5}} & {\large 44.8} & {\large {26.2}} & {\large 72.9} & {\large {42.0}} & {\large 35.5} & {\large 18.3}\\ 
    
    & {\large YOLOX\cite{ge2021yolox}} & {\large RGB}  & {\normalsize {arXiv'2021}}& {\large 76.6} & {\large {\color{blue}\textbf{48.1}}} & {\large 43.5} & {\color{blue}\large {\textbf{26.5}}} & {\large 71.0} & {\large {44.0}} & {\large 27.5} & {\large 17.4}\\
    \hline
    \hline
    
    {\multirow{3}*{{\small \rotatebox{90}{RGB-Event}}}} & {\large FPN-fusion\cite{tomy2022fusing}} & {\normalsize RGB, Voxel\cite{zhu2019unsupervised}}  & {\normalsize {ICRA'2022}}& {\large 56.8} & {\large 30.7} & {\large 34.4} & {\large 18.6} & {\large 63.7} & {\large 36.9} & {\large 36.9} & {\large 19.7} \\
    
    & {\large RENet\cite{zhou2022rgb}} & {\normalsize RGB, E-TMA\cite{zhou2022rgb}}  & {\normalsize {ICRA'2023}}& - & - & - & - & {\large 65.4} & {\large 39.4} & {\large 37.3} & {\large \color{blue} \textbf{22.2}} \\
    
    & {\large SFNet} & {\normalsize RGB, SIF}& - & \color{red}\textbf{{\large 80.0}} & \color{red}\textbf{{\large 50.9}} & \color{red}\textbf{{\large 51.4}} & \color{red}\textbf{{\large 30.4}} & \color{red}\textbf{{\large 75.8}} & \color{red}\textbf{{\large 50.2}} & \color{red}\textbf{{\large 38.0}} & \color{red}\textbf{{\large 24.2}} \\
    \Xhline{1.5pt}
  \end{tabular}
  \label{table2}
  }
\end{table*}

%% file: PDF/Table/Tab_ComparisonWithSotaObjectDet_Exposure.tex
\begin{table*}[tbp]
\belowrulesep=0pt
\aboverulesep=0pt
\centering
  \setlength{\tabcolsep}{1.1mm}
  \caption{Performance comparison with SOTA methods under different exposure conditions (mAP50 / mAP50:95). The best and the second-best performances are
 marked in {\color{red}{\textbf{red bold}}} and {\color{blue}{\textbf{blue bold}}}, respectively.}
  \label{tab:commands}
  \renewcommand\arraystretch{1.9} 
  \scalebox{0.75}{
  \begin{tabular} {ccc|cc|cc|cc|cc}
    \Xhline{1.5pt}
    {\multirow{3}*{{\large Model Type}}} & {\multirow{3}*{{\Large Method}}} & {\multirow{3}*{{\Large Input}}}  & \multicolumn{4}{c|}{\large{DSEC-Det}} & \multicolumn{4}{c}{\large{DSEC-Det-sub}} \\ 
    & &  & \multicolumn{2}{c}{\large{ Class-balanced}} &  \multicolumn{2}{c|}{\large{ Class-imbalanced}} & \multicolumn{2}{c}{\large{ Class-balanced}} &  \multicolumn{2}{c}{\large{ Class-imbalanced}}\\
    \hhline{~~~--------}
    & &  & \normalsize{ daytime} & \normalsize{ nighttime} & \normalsize{ daytime} & \normalsize{ nighttime}  & \normalsize{ daytime} & \normalsize{ nighttime}  & \normalsize{ daytime} & \normalsize{ nighttime}\\
    \hline 
   
    {\multirow{5}*{{\large \rotatebox{90}{RGB}}}} & {\large Faster-RCNN\cite{ren2015faster}} & {\large RGB}  &  {\normalsize {78.9/46.5}} &{\normalsize {61.6/34.1}}  & {\normalsize {47.6/27.0}} & {\normalsize {38.8/20.5}}  & {\normalsize {81.1/50.1}} & {\normalsize {53.5/24.0}} & {\normalsize {50.8/27.3}} & {\normalsize {\color{red}{\textbf{28.8}}/\color{blue}{\textbf{14.6}}}} \\
    
    & {\large RetinaNet\cite{lin2017focal}} & {\large RGB}  &  {\normalsize {65.8/38.2}} &{\normalsize {48.7/26.3}}  & {\normalsize {40.5/23.0}} & {\normalsize {32.0/16.0}}  & {\normalsize {73.6/45.4}} & {\normalsize {41.5/20.7}} & {\normalsize {49.1/26.2}} & {\normalsize {23.9/11.1}} \\
    
    & {\large CenterNet\cite{zhou2019objects}} & {\large RGB}  &  {\normalsize {77.7/47.1}} &{\normalsize {62.5/36.4}}  & {\normalsize {43.1/25.8}} & {\normalsize {36.4/18.3}}  & {\normalsize {79.6/49.8}} & {\normalsize {54.7/26.1}}  & {\normalsize {44.0/23.6}} & {\normalsize {26.5/12.4}} \\
    
    & {\large YOLOv5\cite{Jocher_YOLOv5_by_Ultralytics_2020}} & {\large RGB}  & {\normalsize {81.2/\color{red}{\textbf{54.0}}}} &{\normalsize {62.0/39.9}} &{\normalsize {48.6/\color{blue}{\textbf{31.5}}}} & {\normalsize {36.3/21.7}}  & {\normalsize {77.4/\color{blue}{\textbf{52.2}}}} & {\normalsize {44.8/23.1}}  & {\normalsize {\color{red}{\textbf{54.5}}/\color{blue}{\textbf{30.3}}}} & {\normalsize {19.3/10.3}}\\
    
    & {\large YOLOv7\cite{wang2023yolov7}} & {\large RGB} & {\normalsize {{\color{blue}{\textbf{81.7}}}/50.9}} &{\normalsize {65.2/38.7}}  & {\normalsize {{\color{blue}{\textbf{48.7}}}/30.1}} & {\normalsize {35.8/18.5}}  & {\normalsize {{\color{blue}{\textbf{81.2}}}/52.1}} & {\normalsize {51.2/23.3}} & {\normalsize {47.1/24.4}} & {\normalsize {21.5/10.8}} \\
    
    & {\large YOLOX\cite{ge2021yolox}} & {\large RGB} & {\normalsize {{\color{blue}{\textbf{81.7}}}/52.2}} &{\normalsize {\color{blue}{\textbf{66.2}}/\color{blue}{\textbf{40.8}}}} & {\normalsize {45.9/29.2}} & {\normalsize {\color{blue}{\textbf{42.6}}/\color{blue}{\textbf{24.2}}}}  & {\normalsize {77.7/50.7}} & {\normalsize {\color{blue}{\textbf{57.1}}/\color{blue}{\textbf{26.4}}}} & {\normalsize {36.3/23.0}} & {\normalsize {23.4/12.3}} \\
    \hline
    \hline
    
    {\multirow{3}*{{\normalsize \rotatebox{90}{RGB-Event}}}} & {\large FPN-fusion\cite{tomy2022fusing}} & {\normalsize RGB, Voxel\cite{zhu2019unsupervised}}  &  {\normalsize {63.4/35.8}}  &{\normalsize {38.8/20.3}} & {\normalsize {37.1/20.0}}  &{\normalsize {34.4/18.6}} & {\normalsize {70.2/42.8}} & {\normalsize {43.8/19.9}}  & {\normalsize {44.3/22.9}} & {\normalsize {22.6/10.8}}\\
    
    & {\large RENet\cite{zhou2022rgb}} & {\normalsize RGB, E-TMA\cite{zhou2022rgb}}  & - & - &  - & - & {\normalsize {73.8/45.4}} & {\normalsize {34.5/15.8}}  & {\normalsize {51.1/29.4}} & {\normalsize {17.9/9.0}}\\
    
    & {\large SFNet} & {\normalsize RGB, SIF} & {\normalsize {\color{red}{\textbf{84.4}}/\color{blue}{\textbf{53.9}}}} &{\normalsize {\color{red}{\textbf{71.2}}/\color{red}{\textbf{44.1}}}} & {\normalsize {\color{red}{\textbf{53.1}}/\color{red}{\textbf{33.7}}}} & {\normalsize {\color{red}{\textbf{50.4}}/\color{red}{\textbf{27.9}}}} & {\normalsize {\color{red}{\textbf{83.0}}/\color{red}{\textbf{57.3}}}} & {\normalsize {\color{red}{\textbf{59.5}}/\color{red}{\textbf{32.4}}}} & {\normalsize {\color{blue}{\textbf{53.6}}/\color{red}{\textbf{33.4}}}} & {\normalsize {\color{blue}{\textbf{27.7}}/\color{red}{\textbf{15.2}}}} \\
    \Xhline{1.5pt}
  \end{tabular}
  \label{table22}
  }
\end{table*}

%% file: PDF/Table/Tab_ComparisonWithSotaErAndMc.tex
\begin{table*}[ttbp]
\belowrulesep=0pt
\aboverulesep=0pt
\centering

  \caption{Performance comparison with SOTA Event Representation methods and motion compensation methods. The best and the second-best performances are
 marked in {\color{red}{\textbf{red bold}}} and {\color{blue}{\textbf{blue bold}}}, respectively. MB: Model-based, SSL: self-supervised learning, SL: Supervised learning.}
  \label{tab:commands}
  \renewcommand\arraystretch{2.1} 
  \scalebox{0.65}{
  \begin{tabular} {ccc|cc|cc|cc|cc}
    \Xhline{1.5pt}
    & {\multirow{3}*{{\Large Method}}} & {\multirow{3}*{{\Large Pub. \& Year}}}  & \multicolumn{4}{c|}{{\large DSEC-Det}} & \multicolumn{4}{c}{{\large DSEC-Det-sub}} \\ 

    & &  & \multicolumn{2}{c}{{\large Class-balanced}} &  \multicolumn{2}{c|}{{\large Class-imbalanced}} & \multicolumn{2}{c}{{\large Class-balanced}} &  \multicolumn{2}{c}{{\large Class-imbalanced}}\\
    \hhline{~~~--------}
    & & & {  mAP50} & {  mAP50:95} & {  mAP50}& {  mAP50:95} & {  mAP50} & {  mAP50:95} & {  mAP50}& {  mAP50:95}\\
    \hline
    {\multirow{3}*{{\Large \rotatebox{90}{MB}}}} & {\Large CMax\cite{gallego2018unifying}} & {\Large CVPR'2018} & {\Large 75.5} & {\Large 48.0} &{\Large 50.9}& {\Large 29.1} & {\Large 75.7} & {\Large 49.3} & {\Large 36.0}& {\Large 22.7}\\
    
     &{\Large ST-PPP\cite{gu2021spatio}} & {\Large ICCV'2021} & {\Large 75.3} & {\Large 48.5} &{\Large 49.2}& {\Large \color{blue}{\textbf{30.3}}} & {\Large 73.6} & {\Large 48.6} & {\Large 34.1}& {\Large 21.7}\\
     
     &{\Large MCM\cite{shiba2022secrets}} & {\Large ECCV'2022} & {\Large 79.9} & {\Large 48.9} &{\Large  \color{blue}{\textbf{51.4}}}& {\Large 28.8} & {\Large 75.7} & {\Large 49.3} & {\Large 34.5}& {\Large 22.0}\\
     \hdashline
      {\multirow{2}*{{\Large \rotatebox{90}{SSL}}}} & \thead{\Large{ConvGRU-} \\ \Large{EV-FlowNet\cite{hagenaars2021self}}} & {\Large NeurIPS'2021} & {\Large  \color{red}{\textbf{80.2}}} & {\Large 49.2} &{\Large 47.3}& {\Large 29.6} & {\Large 71.6} & {\Large 48.1} & {\Large 36.7}& {\Large 22.8}\\
      
      &{\Large Federico \emph{et al.}\cite{paredes2023taming}} & {\Large ICCV'2023} & {\Large 76.0} & {\Large 48.5} &{\Large 48.1}& {\Large 29.6} & {\Large 75.0} & {\Large 49.2} & {\Large 37.7}& {\Large 23.9}\\
      \hdashline
      {\multirow{2}*{{\Large \rotatebox{90}{SL}}}} & {\Large E-RAFT\cite{gehrig2021raft}} & {\Large 3DV'2021} & {\Large 79.1} & {\Large 48.9} &{\Large  \color{red}{\textbf{51.9}}}& {\Large  \color{red}{\textbf{30.4}}} & {\Large 73.1} & {\Large 47.2} & {\Large 37.2}& {\Large 22.5}\\
      
      &{\Large IDNet \cite{Wu_2024_ICRA}} & {\Large ICRA'2024} & {\Large 72.8} & {\Large 46.9} &{\Large 47.7}& {\Large 29.2} & {\Large 73.9} & {\Large 47.2} & {\Large \color{red} \textbf{39.1}}& {\Large \color{red}{\textbf{24.3}}}\\

    \hline
    &{\Large Timestamp\cite{park2016performance}} & {\Large ICIP'2016} & {\Large 79.5} & {\Large 49.4} & {\Large 50.6} & {\Large 29.4} & {\Large 73.7} & {\Large 49.2} & {\Large 37.2} & {\Large 22.8}\\
     &{\Large Time Surface\cite{lagorce2016hots}}  & {\Large TPAMI'2016}& {\Large 79.9} & {\Large 49.4} & {\Large 51.3}& {\Large 29.4} & {\Large 75.3} & {\Large 49.4} & {\Large 37.9}& {\Large 23.7}\\
     &{\Large DiST\cite{kim2021n}} & {\Large ICCV'2021} & {\Large 79.4} & {\Large 48.9} & {\Large 50.0}& {\Large 29.4} & {\Large 75.7} & {\Large 49.6} & {\Large 37.7}& {\Large \color{blue}{\textbf {24.2}}}\\
     
     &{\Large Voxel\cite{zhu2019unsupervised}} & {\Large CVPR'2019} & {\Large 79.7} & {\Large 48.8} &{\Large 49.5}& {\Large 28.7} & {\Large 71.1} & {\Large 48.0} & {\Large 35.6}& {\Large 22.8}\\
     
     &{\Large ERGO-12\cite{zubic2023chaos}} & {\Large ICCV'2023} & {\Large  \color{blue}{\textbf{80.1}}} & {\Large  \color{blue}{\textbf{49.7}}} &{\Large 49.1}& {\Large 30.2} & {\Large \color{red}{\textbf{77.4}}} & {\Large  {\color{blue} \textbf{50.1}}} & {\Large 36.7}& {\Large 23.2}\\
     
     &{\Large SIF} & - & {{\Large 80.0}} & \textbf{\color{red}{\Large 50.9}} & \textbf{\color{blue}{\Large 51.4}} & \textbf{\color{red}{\Large 30.4}} & {{\Large \textbf{\color{blue}{75.8}}}} & \textbf{\color{red}{\Large 50.2}} & \textbf{\color{blue}{\Large 38.0}} & \textbf{\color{blue}{\Large 24.2}} \\
    \Xhline{1.5pt}
  \end{tabular}
  \label{table10}
  }
\end{table*}

%% file: PDF/Figure/Fig_QualitativeComparisonWithRGB.tex
\begin{figure*}[htbp]
\centering
\includegraphics[scale=0.19]{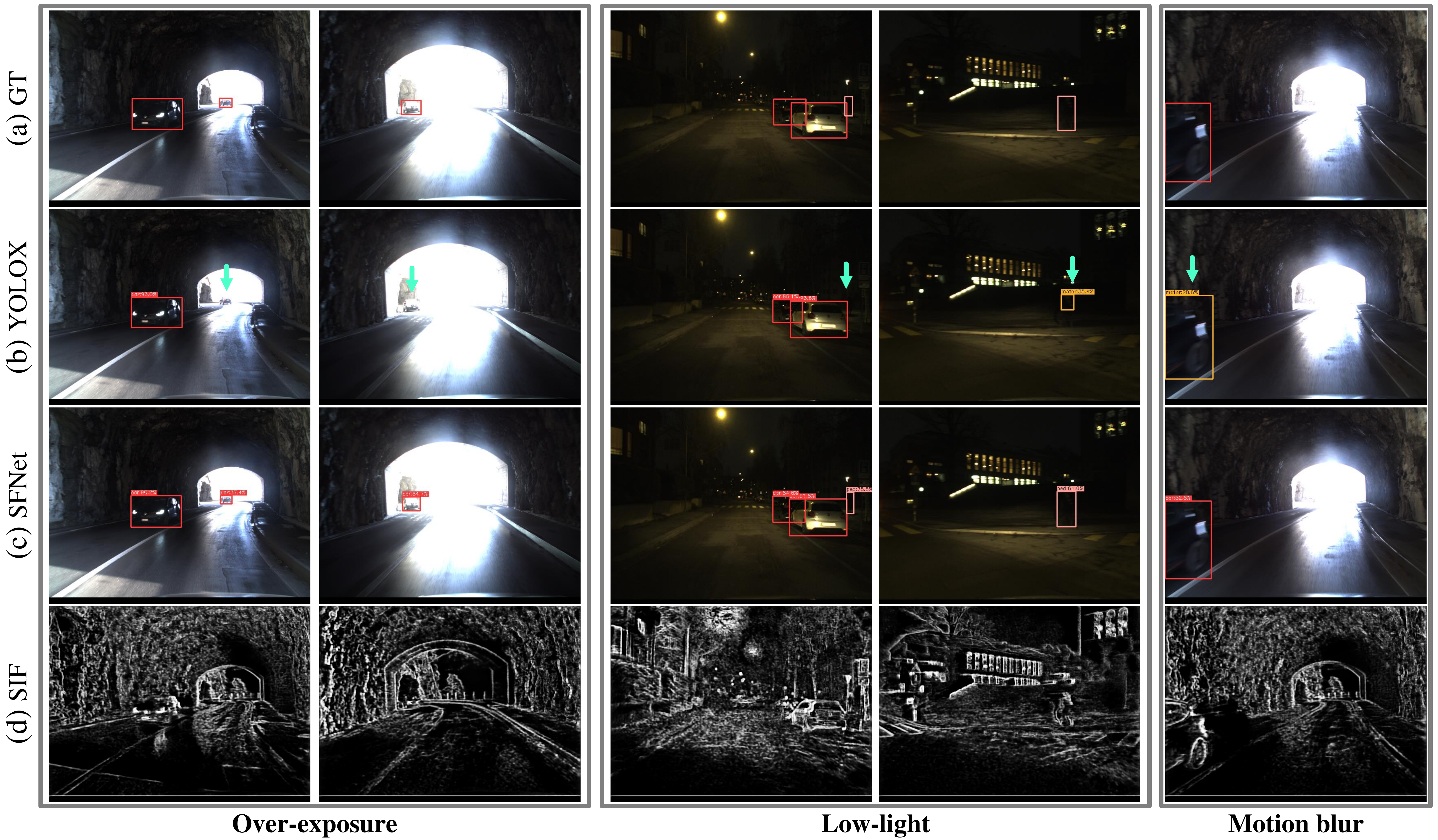}
\caption{Qualitative comparison with RGB baseline on the class-imbalanced situation of the DSEC-Det dataset. We utilize \textcolor[RGB]{11,255,200}{\textbf{green}} arrows to mark the failed cases.}
\label{fig7}
\end{figure*}

%% file: PDF/Figure/Fig_QualitativeComparisonWithEventRepresentation.tex
\begin{figure*}[tbp]
\centering
\includegraphics[scale=0.145]{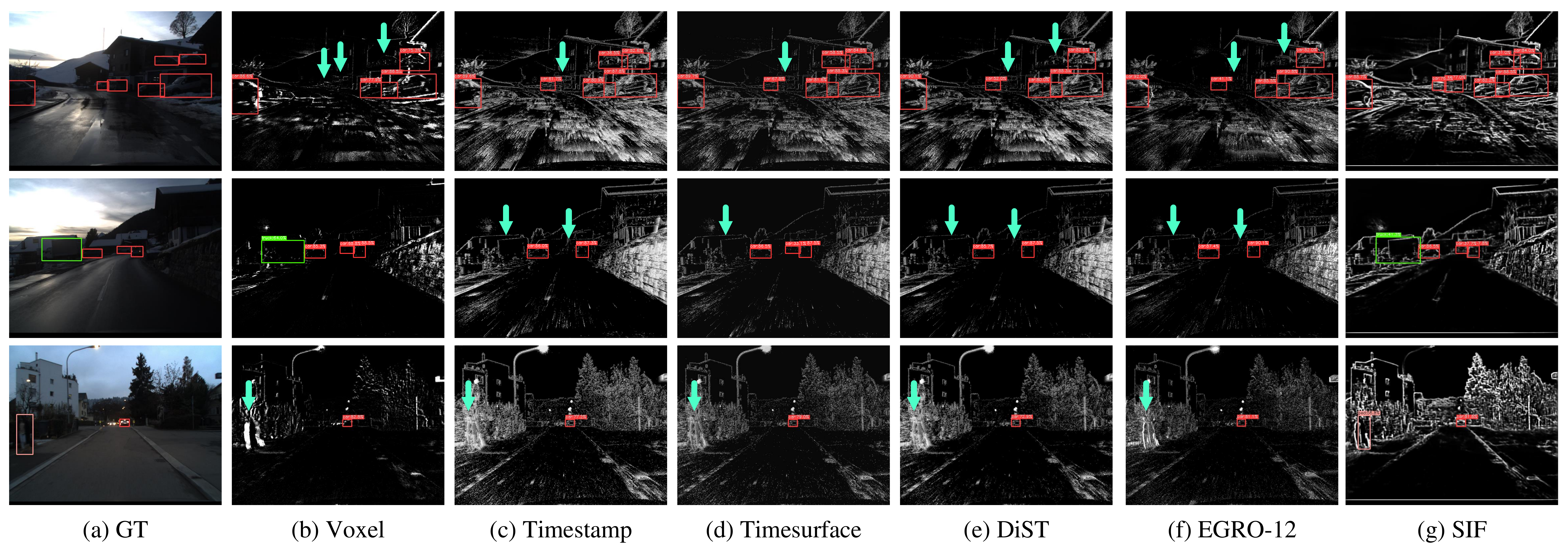}
\caption{Qualitative comparison with SOTA event representation methods on the class-imbalanced situation of the DSEC-Det dataset. We utilize \textcolor[RGB]{11,255,200}{\textbf{green}} arrows to mark the failed cases.}
\label{fig8}
\end{figure*}

%% file: PDF/Figure/Fig_QualitativeComparisonWithMotionCom.tex
\begin{figure*}[tbp]
\centering
\includegraphics[scale=0.101]{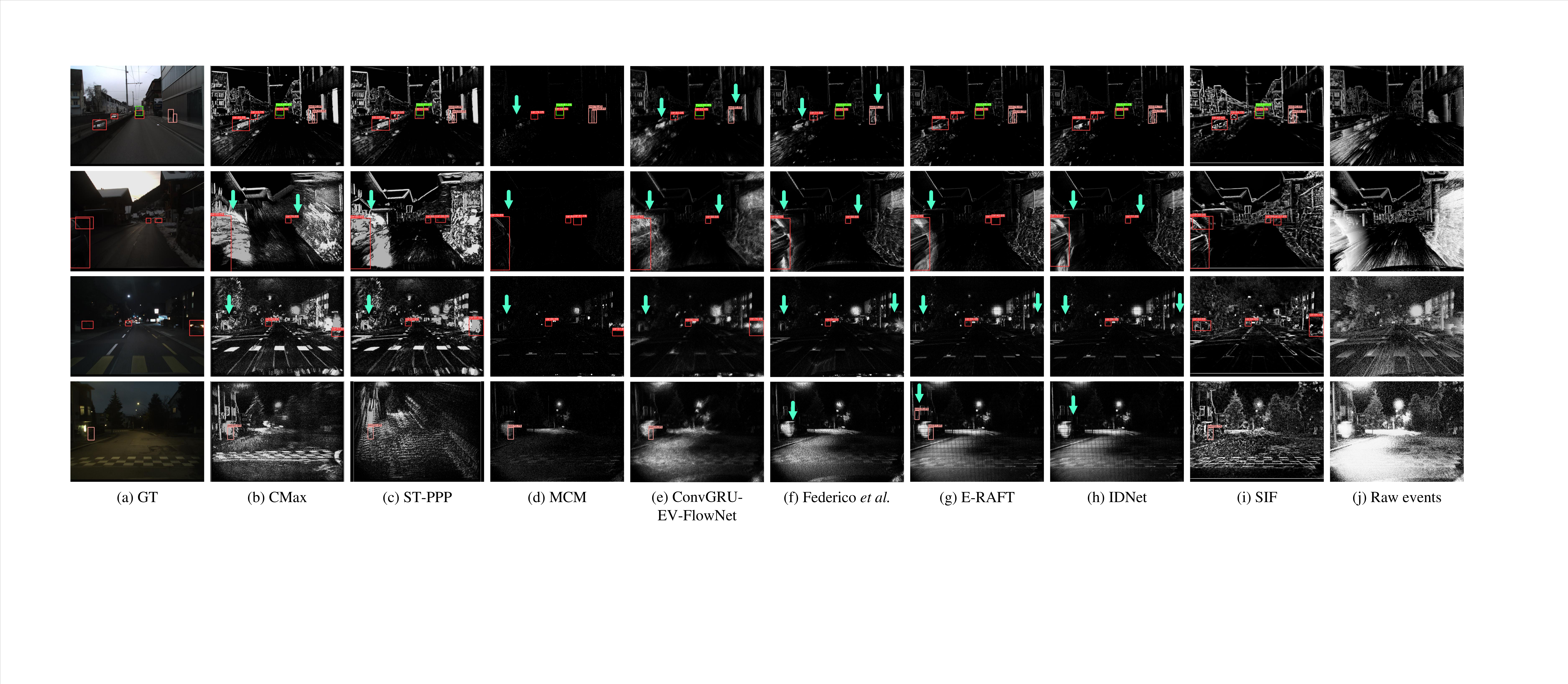}
\caption{Qualitative comparison with SOTA motion compensation methods on the class-imbalanced situation of the DSEC-Det dataset. We utilize \textcolor[RGB]{11,255,200}{\textbf{green}} arrows to mark the failed cases.}
\label{fig11}
\end{figure*}

%% file: PDF/Figure/Fig_VisualizationOfAFCM.tex
\begin{figure*}[tbp]
\centering
\includegraphics[scale=0.16]{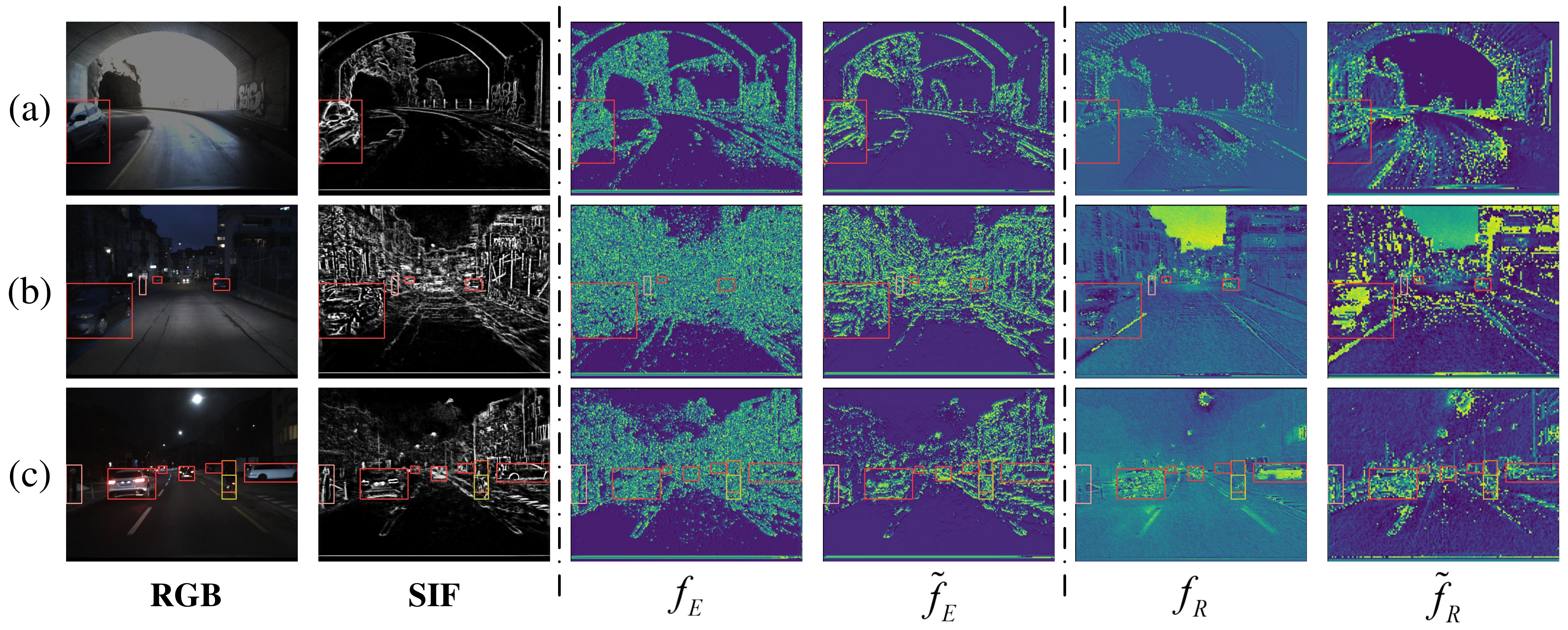}
\caption{Visualization of features before and after AFCM. $f_{E}$ and ${\widetilde{f_{E}}}$ represent the features from the event modality before and after the ERM. $f_{R}$ and ${\widetilde{f_{R}}}$ represent the features from the image modality before and after the LADM.}
\label{fig9}
\end{figure*}

%% file: PDF/Table/Tab_PeformanceOfComponents.tex
\begin{table*}[htbp]
\belowrulesep=0pt
\aboverulesep=0pt
\centering
  \normalsize
  \setlength{\tabcolsep}{1.3mm}
  \caption{THE PERFORMANCE OF OUR COMPONENTS. The \textcolor[RGB]{255,0,0}{\textbf{RED BOLD}} and \textcolor[RGB]{0,0,255}{\textbf{BLUE BOLD}} represent the improvement and slight degradation compared to the baseline. The best performance is marked in \textbf{BLACK BOLD}.}
  \label{tab:commands}
  \scalebox{0.9}{
  \begin{tabular} {ccc|cc|cc|cc|cc}
    \Xhline{1.5pt}
    {\multirow{3}*{SIF}} & {\multirow{3}*{ERM}} & {\multirow{3}*{LDAM}} & \multicolumn{4}{c|}{DSEC-Det} & \multicolumn{4}{c}{DSEC-Det-sub} \\ 

     & & &  \multicolumn{2}{c}{Class-balanced} &  \multicolumn{2}{c|}{Class-imbalanced} & \multicolumn{2}{c}{Class-balanced} &  \multicolumn{2}{c}{Class-imbalanced}\\
    \hhline{~~~--------}
     & & &  mAP50 & mAP50:95 & mAP50& mAP50:95 & mAP50 & mAP50:95 & mAP50& mAP50:95\\
    \hline
     \multicolumn{3}{c|}{YOLOX (RGB)} &  76.6 & 48.1 & 43.5 & 26.6 & 71.0 & 44.0 & 27.5 & 17.4\\ 
     \hdashline
    {\checkmark} &  &   & 79.7\textcolor[RGB]{255,0,0}{{\footnotesize +3.1}} & 47.9\textcolor[RGB]{0,0,255}{{\footnotesize-0.2}} & 49.2\textcolor[RGB]{255,0,0}{{\footnotesize+5.7}} & 28.5\textcolor[RGB]{255,0,0}{{\footnotesize+1.9}} & 75.2\textcolor[RGB]{255,0,0}{{\footnotesize+4.2}} & 43.7\textcolor[RGB]{0,0,255}{{\footnotesize-0.3}} & 36.4\textcolor[RGB]{255,0,0}{{\footnotesize+8.9}} & 21.8\textcolor[RGB]{255,0,0}{{\footnotesize+4.4}}\\ 
    {\checkmark} & {\checkmark} &  & \textbf{80.8}\textcolor[RGB]{255,0,0}{{\textbf{\footnotesize+4.2}}} & 47.7\textcolor[RGB]{0,0,255}{{\footnotesize-0.4}} & 47.9\textcolor[RGB]{255,0,0}{{\footnotesize+4.4}} & 28.0\textcolor[RGB]{255,0,0}{{\footnotesize+1.4}} & \textbf{75.8}\textcolor[RGB]{255,0,0}{{\textbf{\footnotesize+4.8}}} & 49.5\textcolor[RGB]{255,0,0}{{\footnotesize+5.5}} & 36.8\textcolor[RGB]{255,0,0}{{\footnotesize+9.3}} & 23.4\textcolor[RGB]{255,0,0}{{\footnotesize+6.0}}\\ 
    {\checkmark} &  & {\checkmark} & 79.4\textcolor[RGB]{255,0,0}{\footnotesize+2.8} & 48.6\textcolor[RGB]{255,0,0}{\footnotesize+0.5} & 50.3\textcolor[RGB]{255,0,0}{\footnotesize+6.8} & 29.3\textcolor[RGB]{255,0,0}{\footnotesize+2.7} & 75.6\textcolor[RGB]{255,0,0}{\footnotesize+4.6} & \textbf{50.3}\textcolor[RGB]{255,0,0}{\textbf{\footnotesize+6.3}} & 36.1\textcolor[RGB]{255,0,0}{\footnotesize+8.6} & 22.2\textcolor[RGB]{255,0,0}{\footnotesize+4.8}\\ 
    {\checkmark} & {\checkmark}& {\checkmark} & 80.0\textcolor[RGB]{255,0,0}{\footnotesize+3.4} & \textbf{50.9}\textcolor[RGB]{255,0,0}{\textbf{\footnotesize+2.8}} & \textbf{51.4}\textcolor[RGB]{255,0,0}{\textbf{\footnotesize+7.9}} & \textbf{30.4}\textcolor[RGB]{255,0,0}{\textbf{\footnotesize+3.8}} & \textbf{75.8}\textcolor[RGB]{255,0,0}{\textbf{\footnotesize+4.8}} & 50.2\textcolor[RGB]{255,0,0}{\footnotesize+6.2} & \textbf{38.0}\textcolor[RGB]{255,0,0}{\textbf{\footnotesize+10.5}} & \textbf{24.2}\textcolor[RGB]{255,0,0}{\textbf{\footnotesize+6.8}}\\ 
    \Xhline{1.5pt}
  \end{tabular}
  \label{table3}
  }
\end{table*}

%% file: PDF/Table/Tab_PerformaceOfTimeWindow.tex
\begin{table}[tbp]
\belowrulesep=0pt
\aboverulesep=0pt
\centering

  \caption{THE PERFORMANCE OF Different Time Window. The best performance is marked in \textbf{BOLD}.}
  \label{tab:commands}
  \renewcommand\arraystretch{2} 
  \scalebox{0.56}{
  \begin{tabular} {c|cc|cc|cc|cc}

    \Xhline{1.5pt}
    {\multirow{3}*{\large ${T}$}}   & \multicolumn{4}{c|}{{\large DSEC-Det}} & \multicolumn{4}{c}{{\large DSEC-Det-sub}} \\ 

     &   \multicolumn{2}{c}{{\large Class-balanced}} &  \multicolumn{2}{c|}{{\large Class-imbalanced}} & \multicolumn{2}{c}{{\large Class-balanced}} &  \multicolumn{2}{c}{{\large Class-imbalanced}}\\
    \hhline{~--------}
     &  {  mAP50} & {  mAP50:95} & {  mAP50}& {  mAP50:95} & {  mAP50} & {  mAP50:95} & {  mAP50}& {  mAP50:95}\\
    \hline
    {\large 50 ms}   & {\Large 79.4} & {\Large 50.4} & {\Large 50.5} & {\Large 29.6} & {\Large 75.1} & {\Large 49.4} & {\Large 37.9} & {\Large 23.2}\\ 
    {\large 100 ms} & {\Large 80.0} & \textbf{{\Large 50.9}} & \textbf{{\Large 51.4}} & \textbf{{\Large 30.4}} & \textbf{{\Large 75.8}} & \textbf{{\Large 50.2}} & \textbf{{\Large 38.0}} & \textbf{{\Large 24.2}}\\ 
    {\large 150 ms}   & {\Large \textbf{80.3}} & {\Large 50.8} & {\Large 49.8} & {\Large 29.0} & {\Large 73.5} & {\Large 49.7} & {\Large 36.0} & {\Large 22.3}\\ 
    \Xhline{1.5pt}
  \end{tabular}
  \label{table6}
  }
\end{table}

%% file: PDF/Table/Tab_PerformaceOfSuperSignals.tex
\begin{table}[tbp]
\belowrulesep=0pt
\aboverulesep=0pt
\centering

  \normalsize
  \caption{
THE PERFORMANCE OF Different Supervision Signals. The best performance is marked in \textbf{BOLD}.}
  \label{tab:commands}
  \renewcommand\arraystretch{1.5} 
  \scalebox{0.56}{
  \begin{tabular} {c|cc|cc|cc|cc}

    \Xhline{1.5pt}
    {\multirow{3}*{\large ${S}$}}   & \multicolumn{4}{c|}{{\large DSEC-Det}} & \multicolumn{4}{c}{{\large DSEC-Det-sub}} \\ 

     &   \multicolumn{2}{c}{{\large Class-balanced}} &  \multicolumn{2}{c|}{{\large Class-imbalanced}} & \multicolumn{2}{c}{{\large Class-balanced}} &  \multicolumn{2}{c}{{\large Class-imbalanced}}\\
    \hhline{~--------}
     &   mAP50 & mAP50:95 & mAP50& mAP50:95 & mAP50 & mAP50:95 & mAP50& mAP50:95\\
    \hline
    {\large Sobel} & {\Large 80.0} & \textbf{{\Large 50.9}} & \textbf{{\Large 51.4}} & \textbf{{\Large 30.4}} & \textbf{{\Large 75.8}} & \textbf{{\Large 50.2}} & \textbf{{\Large 38.0}} & \textbf{{\Large 24.2}}\\
    {\large Roberts}   & {\Large 79.7} & {\Large 49.9} & {\Large 50.3} & {\Large 29.2} & {\Large 75.4} & {\Large 50.1} & {\Large 36.0} & {\Large 21.6}\\
     {\large Laplace}   & {\textbf{\Large 80.6}} & {\Large 49.9} & {\Large 51.1} & {\Large 28.8} & {\Large 74.3} & {\Large 49.3} & {\Large 35.1} & {\Large 22.1}\\ 
     
    \Xhline{1.5pt}
  \end{tabular}
  \label{table7}
  }
\end{table}

%% file: PDF/Table/Tab_ThePerformanceOfOurModelWithAFCM.tex
\begin{table*}[htbp]
\belowrulesep=0pt
\aboverulesep=0pt
\centering
  \normalsize
  \setlength{\tabcolsep}{1.1mm}
  \caption{The Performance of our model with AFCM placed after different layers. The best performance is marked in \textbf{BOLD}.}
  \label{tab:commands}
  \scalebox{0.9}{
  \begin{tabular} {ccccc|cc|cc|cc|cc}
    \Xhline{1.5pt}
    {\multirow{3}*{Conv}} & {\multirow{3}*{layer1}} &{\multirow{3}*{layer2}}&{\multirow{3}*{layer3}}&{\multirow{3}*{layer4}} & \multicolumn{4}{c|}{DSEC-Det} & \multicolumn{4}{c}{DSEC-Det-sub} \\ 
    &&&&&  \multicolumn{2}{c}{Class-balanced} &  \multicolumn{2}{c|}{Class-imbalanced} & \multicolumn{2}{c}{Class-balanced} &  \multicolumn{2}{c}{Class-imbalanced}\\
    \hhline{~~~~~--------}
     &&&&&  mAP50 & mAP50:95 & mAP50& mAP50:95 & mAP50 & mAP50:95 & mAP50& mAP50:95\\
    \hline
     \multicolumn{5}{c|}{baseline}  & 79.7 & 47.9 & 49.2 & 28.5 & 75.2 & 43.7 & 36.4 & 21.8\\
    \hdashline
     
    {\checkmark} &   &&&  & \textbf{81.5} & 49.5 & 50.0 & 30.3 & 76.1 & \textbf{51.3} & 37.3 & 22.1\\ 
    {\checkmark} & {\checkmark} & &&& 80.0 & \textbf{50.9} & \textbf{51.4} & \textbf{30.4} & 75.8 & 50.2 & 38.0 & \textbf{24.2}\\ 
    {\checkmark} & {\checkmark} & {\checkmark}& && 79.8 & 49.6 & 49.3 & 29.3 & \textbf{77.0} & 48.9 & \textbf{38.8} & 23.6\\ 
    {\checkmark} & {\checkmark} &{\checkmark} & {\checkmark}&& 80.1 & 49.7 & 49.4 & 28.4 & 73.3 & 48.5 & 35.8 & 22.0\\ 
    {\checkmark} & {\checkmark} &{\checkmark} &{\checkmark}&{\checkmark}& 79.8 & 49.7 & 49.4 & 27.1 & 76.3 & 47.8 & 38.3 & 23.0\\ 
    \Xhline{1.5pt}
  \end{tabular}
  \label{table5}
  }
\end{table*}

%% file: PDF/Table/Tab_PerfromaceOfLoss.tex
\begin{table}[tbp]
\belowrulesep=0pt
\aboverulesep=0pt
\centering

  \normalsize
  \caption{The performance of different loss functions for our RSGNet. The best performance is marked in \textbf{BOLD}.}
  \label{tab:commands}
  \renewcommand\arraystretch{1.5} 
  \scalebox{0.53}{
  \begin{tabular} {c|cc|cc|cc|cc}

    \Xhline{1.5pt}
    {\multirow{3}*{{\large $Loss$}}}  & \multicolumn{4}{c|}{{\large DSEC-Det}} & \multicolumn{4}{c}{{\large DSEC-Det-sub}} \\ 
     &   \multicolumn{2}{c}{{\large Class-balanced}} &  \multicolumn{2}{c|}{{\large Class-imbalanced}} & \multicolumn{2}{c}{{\large Class-balanced}} &  \multicolumn{2}{c}{{\large Class-imbalanced}}\\
    \hhline{~--------}
     &   mAP50 & mAP50:95 & mAP50& mAP50:95 & mAP50 & mAP50:95 & mAP50& mAP50:95\\
    \hline
    {\Large ${\phi _{CC}}$}& {\Large 78.8} & {\Large 47.9} & {\Large 50.5} & {\Large 28.6} & {\Large \textbf{76.8}} & {\Large 49.8} & {\Large 37.1} & {\Large 23.0}\\ 
    {\Large ${\phi _{L1}}$+${\phi _{TV}}$}& {\Large 77.9} & {\Large 48.9} & {\Large 47.5} & {\Large 28.4} & {\Large 71.1} & {\Large 48.3} & {\Large 33.9} & {\Large 21.6}\\ 
    
    {\Large ${\phi _{L2}}$+${\phi _{TV}}$}& {\Large 77.2} & {\Large 48.8} & {\Large 46.6} & {\Large 28.8} & {\Large 71.8} & {\Large 47.7} & {\Large 33.7} & {\Large 21.9}\\ 
    
    {\Large ${\phi _{CC}}$+${\phi _{TV}}$} & \textbf{{\Large 80.0}} & \textbf{{\Large 50.9}} & \textbf{{\Large 51.4}} & {\Large \textbf{30.4}} & {\Large 75.8} & {\Large \textbf{50.2}} & \textbf{{\Large 38.0}} & \textbf{{\Large 24.2}}\\ 
    \Xhline{1.5pt}
  \end{tabular}
  \label{table4}
  }
\end{table}

%% file: PDF/Figure/Fig_VisualizationOfRSGNet.tex
\begin{figure}[tbp]
\centering
\includegraphics[scale=0.19]{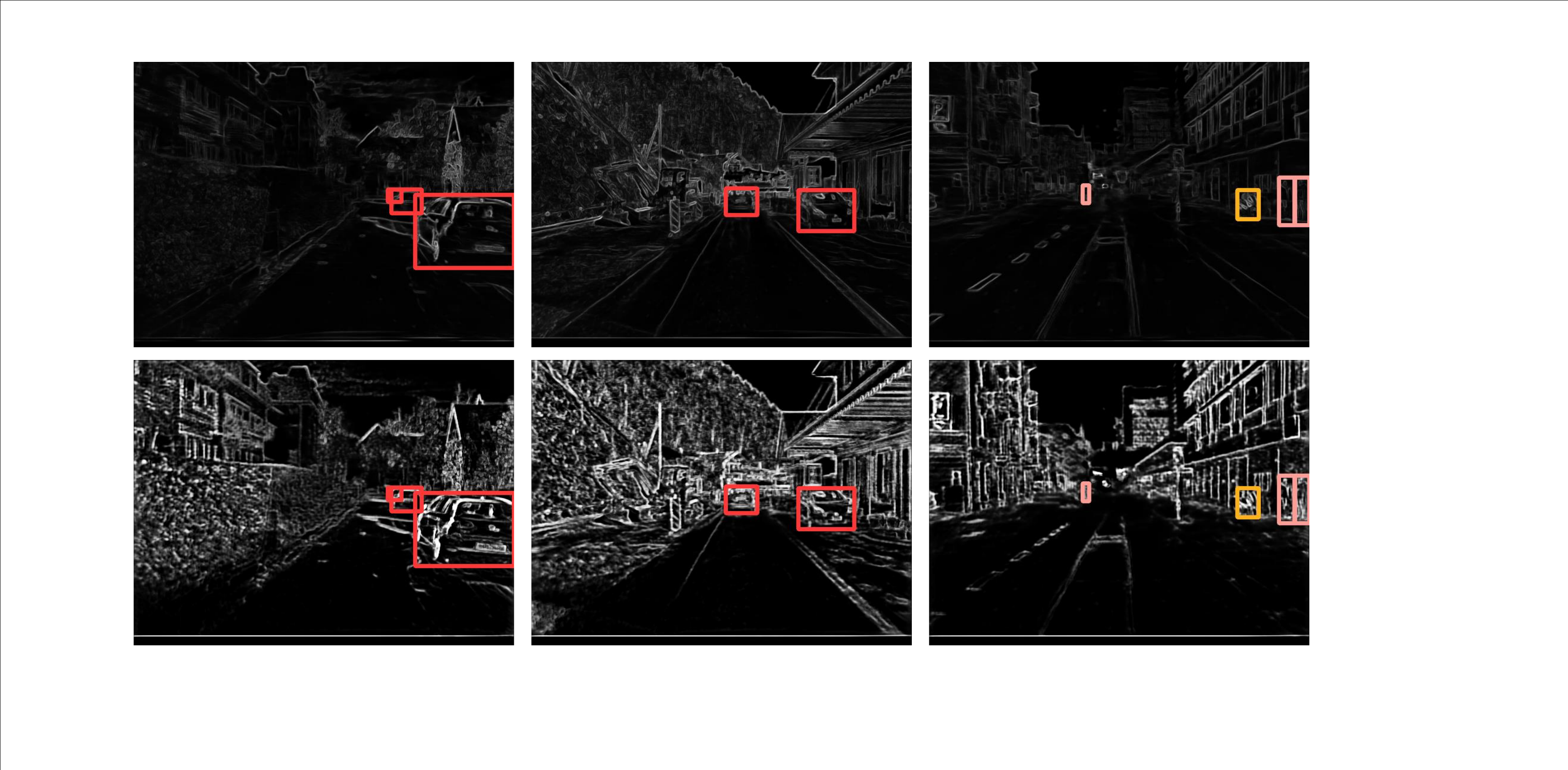}
\caption{The results of RSGNet trained with different loss functions. The first row corresponds to SIF with ${\phi _{CC}}$, while the second row corresponds to SIF with both ${\phi _{CC}}$ and ${\phi _{TV}}$.}
\label{fig10}
\end{figure}